\documentclass{MITcsail}

\definecolor{darkred}{RGB}{140, 21, 21}
\definecolor{citegray}{gray}{0.7}
\definecolor{orange}{HTML}{F58025}

\definecolor{deepred}{rgb}{0.631,0.102,0.102}
\definecolor{amethyst}{rgb}{0.6, 0.4, 0.8}
\definecolor{darkgreen}{rgb}{0.3,0.7,0.3}
\definecolor{salmon}{RGB}{241, 150, 141}
\definecolor{mildyellow}{HTML}{FFF2CC}

\usepackage{xspace}
\usepackage{xcolor}
\usepackage{setspace}

\definecolor{aiblue}{RGB}{66, 133, 244}
\definecolor{humangreen}{RGB}{15, 157, 88}
\definecolor{lightgray}{RGB}{248, 248, 248}
\definecolor{codebg}{RGB}{240, 240, 240}

\newcommand{\airesponsetitle}{}

\newcounter{aimessage}

\newcommand{\humanprompttitle}{}

\newcounter{humanmessage}

\usepackage{enumitem}
\usepackage{cleveref}
\usepackage{float}
\usepackage{array}
\usepackage{makecell}  
\usepackage{booktabs}  
\usepackage{longtable}
\usepackage{multirow}
\newcolumntype{P}[1]{>{\raggedright\arraybackslash}p{#1}}   
\usepackage[T1]{fontenc} 
\usepackage{hyperref}
\usepackage{tikz}
\usepackage[dvipsnames]{xcolor}
\usepackage[most]{tcolorbox}
\usepackage{arydshln}
\usepackage{pifont}

\definecolor{sagegreen}{RGB}{111, 174, 137}
\definecolor{oceanblue}{RGB}{80, 162, 209}
\definecolor{floralpink}{RGB}{209, 146, 207} 

\hypersetup{
    colorlinks=true,
    linkcolor=orange,
    citecolor=gray,
    filecolor=darkred,
    urlcolor=orange,
}

\usepackage{supertabular}

\title{Who Checks the Citations? \\Benchmarking Legal Hallucination Detection}

\author{Patty Liu,$^{\dagger}$
Dominik Stammbach,
Peter Henderson$^{\dagger}$ \\
\normalfont{\small Princeton University}
}
\vspace{1em}

\begin{document}

\maketitle
\thispagestyle{firstpagestyle}

\setcounter{footnote}{2}
\renewcommand\thefootnote{\fnsymbol{footnote}}                                
\footnotetext{Correspondence to: patty.liu@princeton.edu; peter.henderson@princeton.edu.}                                          
  \setcounter{footnote}{0}                                                      
\renewcommand\thefootnote{\arabic{footnote}}

\noindent\href{https://huggingface.co/datasets/ai-law-society-lab/Legal_Phantom_Citation}{Dataset} | 
\href{https://github.com/princeton-polaris-lab/legal-hallucination-agent}{Code} | \href{https://princeton-polaris-lab.github.io/legal-hallucination-webpage/}{Website}
\newcommand{\draftComment}[1]{#1}
\newcommand\patty[1]{\draftComment{{\color{teal}[\textit{#1}]$_{-PL}$}}}
\newcommand\jonas[1]{\draftComment{{\color{violet}[\textit{#1}]$_{-JG}$}}}
\newcommand\dominik[1]{\draftComment{{\color{orange!75!black}[\textit{#1}]$_{-DS}$}}}
\newcommand\inyoung[1]{\draftComment{{\color{blue}[\textit{#1}]$_{-IY}$}}}

\begin{abstract}

Attorneys, judges, and pro se filers increasingly use AI to draft legal documents, yet these tools frequently fabricate citations. Despite predictions that newer models would hallucinate less or that court sanctions would deter negligent filers, we found over 1{,}000 filings containing fabricated citations---with this number growing year-over-year. 
This study evaluates whether AI-based systems can mitigate these errors by automatically detecting hallucinations. 
We propose a taxonomy of legal citation hallucinations grounded in actual court filings and introduce a dataset of 1{,}300 brief excerpts containing injected errors.
Benchmarking five models in agentic and non-agentic settings reveals that while the latest iterations perform better—GPT-5 achieves 84.4\% recall and a 55.0\% F1 score in an agentic framework—all models struggle with subtle error categories. Agentic verification remains resource-intensive, with GPT-5 averaging 15.3 steps per excerpt. Furthermore, restricted information access limits the efficacy of even the best agents. This gap creates policy concerns, as it disadvantages both AI systems and litigants who lack subscriptions to commercial legal databases. Together, our dataset, tools, and policy recommendations provide a foundation for building and auditing reliable legal citation checking tools.

\end{abstract}

\section{Introduction}\label{sec:introduction}

\begin{figure}[h]
  \centering
\includegraphics[width=0.6\linewidth]{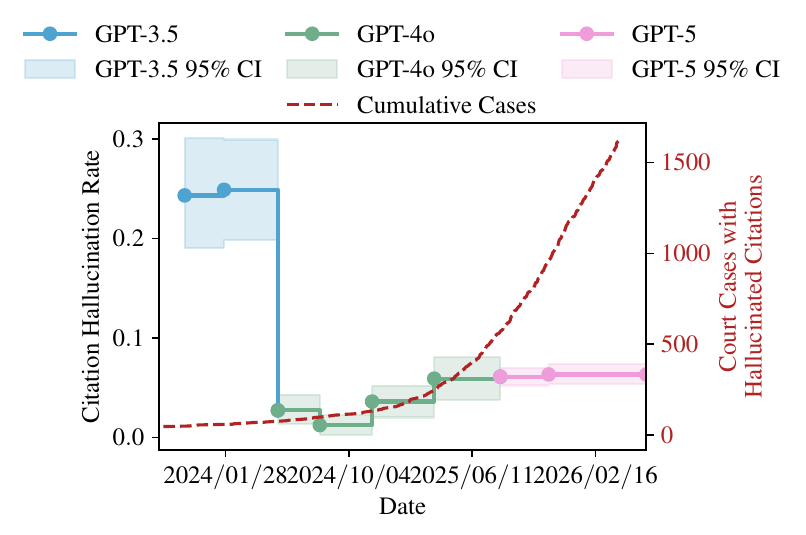}
  \caption{\textbf{Hallucination trend in legal filings and model hallucination rates over time.} Legal hallucination rates are not consistently decreasing across GPT models, while hallucinated citations in court filings are steadily growing. Hallucination rate is the percentage of hallucinated citations in all model generated citations (See \S~\ref{sec:background} for more details).}
  \label{fig:hallucination_trends}
\end{figure}

Legal citations play a prominent role in U.S. legal practice. Attorneys must point to past judicial decisions and laws to make their case~\citep{Duxbury_2008, sep-legal-reas-prec}.
Fabricating citations, or misrepresenting the content of those citations, is the same as pointing to made-up law to win the case. This risks the integrity of the judicial process---courts have called it ``an abuse of the adversary system''. \cite{mata2023avianca}; \textit{see also} \cite{park2024kim} (quoting \textit{Mata}); \cite{noland2025land} (quoting \textit{Mata}).

In the past, such fabrications existed but were rare.\footnote{\emph{See, e.g.,} \textit{Gonzalez-Ayala v. United States}, No. 3:05-cv-01291 (D.P.R. Dec. 2, 2009), ECF No. 8 (discussing such a made-up citation).}
The rapid adoption of large language models (LLMs) in the legal system, though, has turned individual instances into a systemic problem. Pro se litigants,\footnote{Litigants who represent themselves without an attorney.} trained attorneys, and even judges are using LLMs to generate briefs, motions, and other court filings.
But even the best LLMs still fabricate or misrepresent legal precedent at a non-negligible rate (See Figure \ref{fig:hallucination_trends} and \S~\ref{sec:background}).
Combining our own research with other public datasets, we identify over 1{,}000 court filings containing hallucinated citations.\footnote{See \href{https://princeton-polaris-lab.github.io/legal-hallucination-webpage/}{project website} for our compilation of tracked cases, which is merged with additional data from \citet{damientracker}.} Judges repeatedly describe the resulting burden as an ``enormous waste of judicial resources,''
and increasingly impose sanctions because ``lesser sanctions have been insufficient to deter the conduct.'' \cite{powhatan2024skinger, mid025hoosiervac}.

A prevailing argument has been that this problem is temporary, that hallucination rates would diminish as models improve, and that high-profile sanctions would induce greater caution. Our findings and recent events challenge these assumptions. 
Real-world court filings containing hallucinated citations are growing steadily, and the problem is not yet self-correcting (see Figure \ref{fig:hallucination_trends}). Through a controlled experiment querying eight generations of ChatGPT models on 92 legal drafting prompts, we find that hallucination rates are no longer consistently decreasing across model generations, with GPT-5.1 producing hallucinated citations at a significantly higher rate than the best mid-2024 GPT-4o release ($p = 0.001$). Even if hallucination rates improved, the verification burden would continue to grow. Newer models generate more citations per document than their predecessors, drawn from a broader and less canonical set of cases that are individually harder to verify (Figure \ref{fig:citation_in_generated}). The result mirrors the dynamic that \citet{chang2026jevons} calls 
legal practice's own Jevons paradox: as AI reduces the per-unit cost of legal drafting, total filing volume is likely to grow, increasing verification burdens even if individual hallucination rates 
improve.\footnote{See 
\cite{jevons1865coal, abajevons2025, artificiallawyer2024jevons} for more information on Jevons paradox and its application to legal AI.}

We argue that automated verification tools, public access to legal data, and targeted AI literacy guidance for unrepresented filers offer structural paths forward (\S~\ref{sec:discussion}). This work focuses on the first: currently, verifying case citations is largely a manual process performed by attorneys, law clerks, or judges. Especially when a cited case does not exist, this can take significant time and resources. Yet no prior benchmark evaluates how well AI systems can verify legal citations---instead, most research focuses on whether LLMs produce factually correct legal explanations, case summaries, or answers to legal questions
\citep{deroy2023readypretrainedabstractivemodels,savelka2023explaininglegalconceptsaugmented, 2021improving,Dahl_2024,hu-etal-2025-fine,fan2026halluhardhardmultiturnhallucination} (\S~\ref{sec:related}). 

To assess the promise of AI in automatically checking legal filings, we introduce \textsc{LePhantomCite},\footnote{Short for Legal Phantom Citation.} a new benchmarking dataset of legal brief excerpts augmented with injected hallucinations. The dataset enables us to benchmark citation verification systems and is           
  accompanied by a taxonomy of legal citation hallucinations derived from failure modes observed in real court filings.
Using this dataset, we evaluate several LLMs and an agentic system based on \cite{zheng2026learning} that integrates search tools and structured reasoning. Our best agent reliably detects non-existent cases and case name mismatches, but
struggles with verifying pincites,\footnote{A pincite specifies the exact page or paragraph within a source where the cited material appears.} misquotes, and content misrepresentations. The agent also surfaced several citation errors in pre-LLM briefs submitted to state supreme courts, suggesting practical value beyond the benchmark. 
  
Taken together, we make the following contributions:                     
\begin{itemize}[nosep]                                                              
\item We provide longitudinal empirical evidence, across eight ChatGPT models from late 2023 through late 2025, that legal citation hallucination rates do not consistently decline, while the number of hallucinated court filings grows steadily.
\item We introduce a dataset of 1{,}300 legal brief excerpts with injected hallucinations to support evaluation and auditing of citation verification systems, grounded in a taxonomy of legal citation hallucinations derived  
from real court filings.                                                                
\item We evaluate several different models using a custom harness, including access to legal database searches and information-directed exploration. Agentic retrieval improves recall by 26.2\% over non-agentic baselines on GPT-5, but all models struggle most with  
incorrect pincites, verbatim misquotes, and content misrepresentation (52.8\%, 95.2\%, and 83.2\% recall for GPT-5 respectively).        
\item We identify potential policy options for improving the state of automated agentic citation checking. Some failures were due to lack of easy access to publicly available data on legal citations or page number information, making it especially challenging to verify this information. Improving data accessibility for the general public would, in turn, reduce burdens on courts.
\end{itemize} 

Overall, we hope to incentivize improvements and further research in automated verification to ease increasing burdens on the legal system. Our benchmarking effort presents a first step toward that goal.
\section{Hallucination Rate Trends and the Growing Verification Burden}\label{sec:background}

To demonstrate the pervasive challenge of hallucinations in legal content, we conduct a controlled experiment to assess whether newer and better models have reduced citation hallucination rates. We query eight generations of ChatGPT models,\footnote{All eight models were the default model for free-tier ChatGPT access in the browser interface at the respective times, likely to be used by pro se litigants and even many attorneys.} spanning from the first GPT-3.5 model through GPT-5.1, on the same 92 prompts designed to generate legal documents. The prompts ask the LLM to generate documents similar to the ones flagged in real court filings because of AI-generated hallucinations (examples in Appendix Table \ref{tab:legal_doc_prompt}), producing over 8{,}000 case citations in total. We verify all citations using CourtListener and Westlaw.\footnote{CourtListener is a free, open-access repository of U.S. court opinions maintained by the nonprofit Free Law Project and Westlaw is a subscription-based legal research platform by Thomson Reuters.} We classify a citation as hallucinated if it does not correspond to a real case or if the generated case name does not match the name associated with the reporter reference in Westlaw. For example, the citation \textit{995 F.2d 348} in \textit{"In re Grand Jury Subpoenas, 995 F.2d 348, 352–53 (2d Cir. 1993)"} exists, but refers to the case \textit{Malcolm v. National Gypsum Co.} We classify it as hallucinated due to the inconsistency. This definition captures only the most obvious hallucinations and provides a conservative lower bound on overall prevalence of such hallucinations. Subtler misrepresentations, such as citing a real case for a non-supported proposition, are not captured by this criterion.

Figure~\ref{fig:hallucination_trends} shows the cumulative number of identified court filings containing hallucinated citations and hallucination rates observed across models in our experiment. Early GPT-4o models released in mid-2024 exhibit the lowest hallucination rates at 1.23\%, substantially improving over citation rates of around 25\% in earlier GPT-3.5 models.
However, this trend does not persist and more recent models hallucinate legal citations more often again. The GPT-5.1 model generates hallucinated citations at 6.57\%, a higher rate than the August 2024 GPT-4o release (using bootstrapping, p$p = 0.001$).

\begin{figure}
    \centering
    \includegraphics[width=0.9\linewidth]{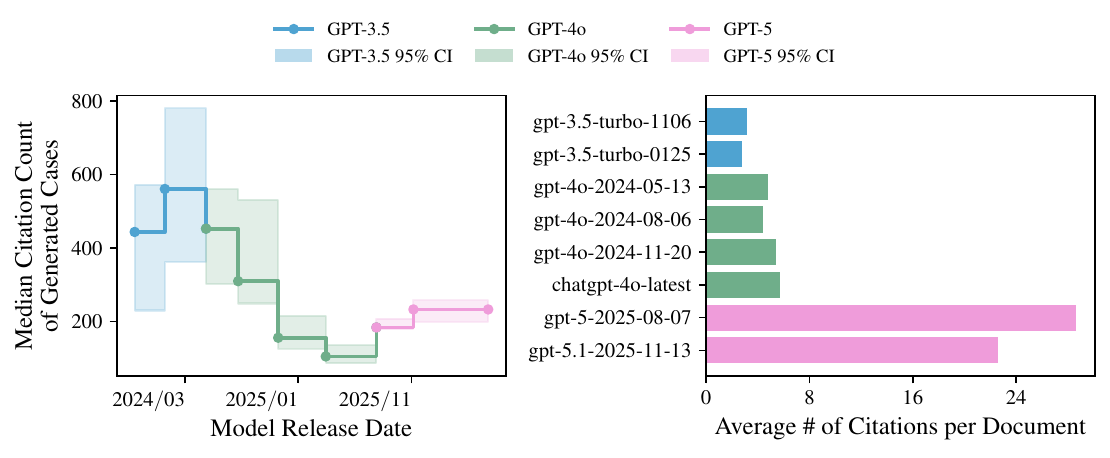}
    \caption{\textbf{Citation counts of the generated cases (left) and average number of citations per document (right).} GPT-4o models generate citations to the most diverse set of cases, with a median citation count of 105 per generated case (left), whereas ChatGPT3.5 generated case citations to mostly widely known landmark cases with a median citation count of over 500. Newer models produce more legal citations per document (right).}\label{fig:citation_in_generated}
\end{figure}

The hallucination rate alone does not fully capture the growing verification burden. Newer models also generate substantially more citations per document than their predecessors, and draw these citations from a broader set of cases: GPT-3.5 models mostly cite well-known landmark cases with a median citation count of over 500, compared to GPT-4o citing less canonical cases with a median citation count of 105 at the lowest. While greater citation diversity may produce better and more tailored legal arguments, lesser-known cases are harder to verify (See Figure \ref{fig:citation_in_generated}).

The citation verification burden affects not just filers. When filers do not verify citations, the task shifts to opposing counsel, law clerks, and judges, whom courts agree are ill-positioned to handle it. See, e.g., \cite{jakes2025youngblood, bischoff2025dept, fang2025hechalou, nelson2025hnavient, takefman2025pickleball, lee2025rr}. Attorneys already cite the lack of time as a reason for not carefully checking AI-generated citations in their own filings. See, e.g., \cite{dehghani2025castro, kaur2025desso, lexos2026overstock}. 

The problem especially impacts pro se litigants, who courts have recognized stand to benefit the most from AI's ability to improve access to justice. \cite{inrebryant2025}. Yet they are also the least equipped to detect hallucinated citations. Courts have observed that AI tools ``increase [both] pro se litigants' ability to draft voluminous pleadings and increase the odds that such pleadings contain [...] 'hallucinated' legal propositions." \cite{mitchel2025stellantis}. Unrepresented litigants lack access to commercial legal databases and are often unaware that AI can fabricate case references. \cite{ligeri2025amazon, alhamin2024star}. Courts have responded to these failures with escalating sanctions, finding that earlier, lighter interventions were insufficient to deter the conduct. \cite{mid025hoosiervac}. But sanctions are reactive and unevenly applied, and do not address the structural conditions driving the problem (See \S~\ref{sec:discussion} for more details).

\section{Methods}

In \S~\ref{sec:background}, we have shown that hallucination rates are not consistently decreasing, and even if they were, verification burdens would increase nonetheless.
Thus, we investigate whether AI can help reduce this burden through automated verification. Because existing datasets do not support evaluation of legal citation verification, we construct the new \textsc{LePhantomCite} dataset. Based on a taxonomy derived from court cases discussing hallucinated citations, we inject different hallucinated citations in the existing legal brief excerpts.

\subsection{Taxonomy of Legal Citation Hallucinations}
\label{sec:taxonomy}

\textbf{Non-existent citation.} 
The citation does not correspond to any real case. In \cite{sims2025souily}, the pro se plaintiff cited "\textit{Graham v. Nyquist} (1974)", but the citation does not point to any existing case. We generate non-existent citations by altering the reporter volume, reporter abbreviation, or page number to implausible values.
    
\textbf{Case name mismatch.} 
The reporter citation and case name refer to two different real cases. In \cite{bevins2025colgate}, the plaintiff's attorney cited "\textit{Tinch v. Video Indus. Servs., Inc.}, No. 1712 EDA 2018, 2019 WL 1396975, at *3 (Pa. Super. Ct. Mar. 27, 2019)", but the Westlaw citation "2019 WL 1396975" corresponds to \textit{Dolberry v. Jakob}, 2019 WL 1396975 (N.D.N.Y. Mar. 28, 2019) instead.
We create mismatches by replacing either the case name or the reporter citation with that of another existing case's. 

\textbf{Incorrect pincite.} The citation refers to the correct case, but the cited page number does not support the quoted language or proposition. In \cite{sasc2024school}, the attorney did not provide a pincite and was warned to provide pinpoint citations in future filings. The court stated that they will "disregard citations to caselaw that are not supported by pinpoint citations," highlighting the importance of providing correct pincites.
    
\textbf{Verbatim misquote.} The exact quoted language does not appear in the cited case. In \cite{harris2025take}, the plaintiff cited "\textit{Chambers v. NASCO, Inc.}, 501 U.S. 32 (1991)" in their motion with a quotation that “courts have ‘the inherent power to police themselves and to sanction bad-faith litigation conduct.” This quote was not found in the cited decision.
We generate verbatim misquotes by replacing one or two words in the original quotation with semantically similar synonyms. This preserves the meaning of the sentence, so there exists no content mismatch (which we define as a separate category below).
    
\textbf{Content misrepresentation.} A cited case exists but does not support the proposition for which it is cited. In \cite{jakes2025youngblood2}, the defendant's attorney wrote, "Moreover, the Pennsylvania Superior Court in \textit{Blackwell v. Eskin} emphasized that even where statements are embarrassing or upsetting, the Plaintiff must demonstrate their precise defamatory content and origin. [...] 916 A.2d 1123, 1128–29 (Pa.
Super. Ct. 2007)." However, the case does not contain any opinion related to identifying defamatory words of the speaker in a complaint.
We generate misrepresentation by altering the holdings so that their meanings change and are incorrect. To add realistic LLM-generated content misrepresentations, we additionally include examples from the Large Legal Fiction dataset \citep{Dahl_2024}.

These categories are not mutually exclusive: a citation may contain multiple hallucination types (e.g., a mismatched case name and a misquote). However, for dataset construction and evaluation, we only introduce a single type of hallucination to a given citation.

\begin{table}[t]
  \centering
    \small

  \begin{tabular}{p{4cm}|p{3.5cm}|p{7cm}}
    \toprule
    \textbf{Hallucination type} & \textbf{Modified component} & \textbf{Example} \\
    \midrule
    Non-existent citation & reporter citation &
    Original: \textcolor{Green}{133 S.\,Ct.\ 1017} \newline $\rightarrow$ Hallucinated: \textcolor{Red}{446 Cal.\ Rptr.\ 4th 183}
    \\
    \hline
    Mismatched case name & case name or reporter citation & 
    Original: \textcolor{Green}{Cinel v. Connick}, 15 F.3d 1338 \newline $\rightarrow$ 
    Hallucinated: \textcolor{Red}{Boone v. Vinson}, 15 F.3d 1338 \\
    \hline
    Incorrect pincite & reporter citation, pincite &
    Original: 830 F.3d at \textcolor{Green}{514} \newline $\rightarrow$ Hallucinated: 830 F.3d at \textcolor{Red}{511}\\
    \hline
    Verbatim misquote & quote &
    Original: ``[T]he parties’ dispute over arbitrability specifically \textcolor{Green}{falls} within those \textcolor{Green}{carve-outs}.'' \newline $\rightarrow$ Hallucinated: ``[T]he parties’ dispute over arbitrability specifically \textcolor{Red}{resides} within those \textcolor{Red}{exclusions}.'' \\
    \hline
    Content misrepresentation & holding &
    Original: \textcolor{Green}{These rules cannot support a claim for retaliatory discharge.} \newline $\rightarrow$ Hallucinated: \textcolor{Red}{Any private policies cannot support a claim for retaliatory discharge under Kansas law.}
    \\
    \bottomrule
  \end{tabular}
  \caption{\textbf{Taxonomy of legal citation hallucinations.} We categorize legal citation hallucinations into five broad categories and show them with examples and relative verification difficulty.}
  \label{tab:taxonomy}
\end{table}

\subsection{LePhantomCite Benchmarking Dataset}
\label{sec:dataset}

The \textsc{LePhantomCite} benchmarking dataset contains 1{,}300 entries drawn from two complementary sources: (1) 1{,}000 excerpts from real appellate briefs with systematically injected hallucinations, covering all five hallucination types introduced in the previous section; and (2) 300 entries from the LLM-generated central holdings portion of \cite{Dahl_2024}, which we manually verify and reformat. We combine these sources because \cite{{Dahl_2024}} provides real LLM-generated examples of content misrepresentation. Across the full dataset, there are 4,499 total citation instances, of which 1,107 citation instances contain hallucination. Figure \ref{fig:dataset_stats} in the Appendix summarizes the count per hallucination type.

\textbf{Appellate brief excerpts.} We collect 245 federal appellate briefs filed in 13 U.S. Courts of Appeals between January 2012 and December 2021, retrieved via the CourtListener API. We restrict to pre-2022 filings to minimize the risk that the source documents themselves contain AI-generated content, and select briefs submitted to Courts of Appeals to ensure high baseline citation quality.
Briefs are converted from PDF to plain text using olmOCR \citep{olmocrbench}, parsed into sentences using a fine-tuned RoBERTa sentence segmentation model \citep{sanchez-2019-sentence}, and then grouped into semantically coherent segments using \texttt{Llama-3.3-70B-Instruct} \citep{grattafiori2024llama3herdmodels}. This yields 5,648 segments; we subsample 1,000, excluding table-of-contents sections and segments with no case citations.

For each citation in the sampled segments, we use \texttt{Qwen3-32B} \citep{yang2025qwen3technicalreport} to extract the associated case name, any quotation, and the holding proposition, constrained to exact substrings of the original text to prevent model-introduced errors during extraction.
We then inject hallucinations by systematically modifying one or more citation components according to the taxonomy. Hallucinations are introduced into 50\% of segments (500 excerpts). We choose this ratio to create a balanced evaluation set, which does not reflect the real hallucination rate observed in generated legal documents.

\paragraph{Hallucination injection.} See Appendix \ref{app:injection_mechanism} for details on hallucination injection mechanism for each type. Non-existent and case name mismatch hallucination types are applied globally: when a citation is replaced with a non-existent citation or a mismatched case name, all instances of it throughout a brief are altered consistently. This prevents a verification model from detecting the hallucination by cross-referencing correct occurrences elsewhere.

\paragraph{LLM-generated holdings.}
Content misrepresentation is the hardest hallucination type to inject synthetically. Thus, we supplement the dataset with 300 entries from the central holding task of \citet{Dahl_2024}, in which LLMs are prompted to state the primary holding of a case given its reporter citation and year. The original hallucination labels in that dataset are based on self-consistency between two LLM outputs rather than ground truth verification. We manually verify all 300 entries against Westlaw, correcting labels where necessary. This process results in 42 confirmed non-hallucinated entries out of 300. See Appendix \ref{app:dataset} for more details on dataset construction. All holding dataset entries are verified using CourtListener and Westlaw.

\subsection{Models and Agent Harnesses }

We evaluate five models with an agentic verification system built on the Bayesian Optimal Experimental Design (BOED) agent framework similar to the frameworks in \cite{choudhury2026bedllmintelligentinformationgathering, lidayan2026abbellearningnaturallanguagebelief, murphy2026agenticforecastingusingsequential}.\footnote{The agent harness is originally based on \href{https://github.com/princeton-polaris-lab/strategic-exploration}{this project}. For all practical purposes this is functionally similar to other cited works' agent harnesses. We also provide full prompts and code \href{https://github.com/princeton-polaris-lab/legal-hallucination-agent}{here}.} In BOED, the agent maintains an explicit, language-based belief state that is updated after each action. We adopt this framework because case citation verification is inherently sequential and information-dependent: the agent must extract citations from the brief excerpts, decide on how to gather information and when it has gathered sufficient evidence to make a hallucination determination. BOED's information-directed search loop is well-suited to this structure. 
We also experiment with the Reflexion agent framework from \cite{shinn2023reflexionlanguageagentsverbal}, but Reflexion agent struggles to keep track of its current decisions on citation validity when the action sequence grows long. The belief state in BOED solves this issue.

Concretely, the agent's belief state is a language-based record that tracks: (1) all case citations, quotations, and holdings extracted from the input excerpt, and (2) the agent's current assessment: correct, hallucinated, or pending, for each element. The belief state is updated after every action and serves as working memory across the multi-step verification process, which is necessary because legal excerpts can contain many citations, each requiring independent lookup chains. The agent has access to eight actions, including legal database search and web search (See Appendix Table \ref{tab:actions} for list of all actions and corresponding descriptions).
\section{Experimental Results}
\subsection{Experimental Setup}
\textbf{Datasets.} 
We divide the dataset into train and test with a 70-30 split. We run all evaluations on the test set, which contains 390 examples. 

\begin{figure}[tbp]
    \centering

    \begin{minipage}[t]{0.55\linewidth}
        \vspace{0pt}
        \centering
        \includegraphics[width=\linewidth]{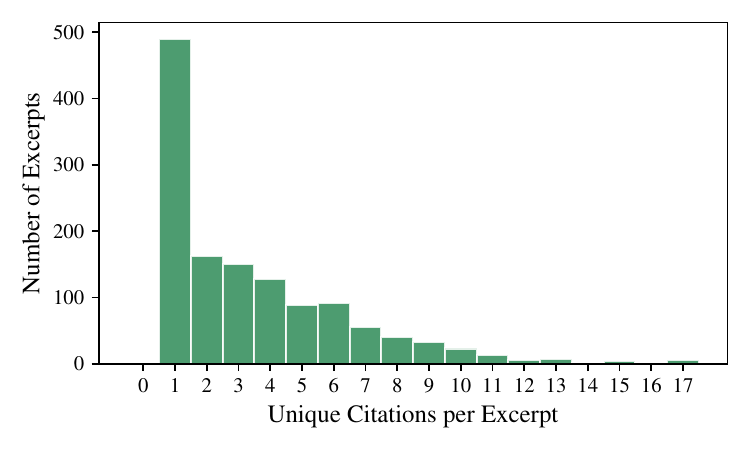}
    \end{minipage}
    \hfill
    \begin{minipage}[t]{0.4\linewidth}
        \vspace{0pt}
        \centering
        \small
        \begin{tabular}{lr}
            \toprule
            \textbf{Hallucination type} & \textbf{Count} \\
            \midrule
            Non-existent citation   & 158 \\
            \hline
            Mismatched case name    &  189  \\
            \hline
            Incorrect pincite       & 177    \\
            \hline
            Verbatim misquote & 167 \\
            \hline
            Content misrepresentation & 416 \\
            \bottomrule
            Total & 1{,}107 \\
            \bottomrule
        \end{tabular}
    \end{minipage}

    \caption{\textbf{Legal citation verification dataset summary.} Distribution of number of unique citations in each brief excerpt (left) and the number hallucinated citations belonging to each hallucination type (right).}
    \label{fig:dataset_stats}
\end{figure}

\textbf{Baselines.}
The agentic framework is model-agnostic. We evaluate it with several open-source and closed-source LLMs, including GPT-5 (gpt-5-2025-08-07) \citep{singh2026openaigpt5card}, GPT-OSS-120B \citep{openai2025gptoss120bgptoss20bmodel}, Qwen3-8B \citep{yang2025qwen3technicalreport}, Qwen3.6-27B, and Gemini-2.5-flash \citep{comanici2025gemini25pushingfrontier}. 
We also evaluate non-agentic performance of these models. 
Experiments using the agentic framework all use $max\_steps=30$ per episode (one agent run on a single excerpt). As shown in Figure \ref{fig:dataset_stats} (a), all excerpts contain less than 18 case citations, with a median of 2 citations per excerpt. Some case citations also have associated quotations and holdings, thus we believe 30 steps to be an adequate number to verify all citations within an excerpt. Non-agentic baselines use the same prompt as the agent but without the belief update and action selection sections. We show the full agentic prompt in Appendix \ref{app:model_prompts}. 

\textbf{Claude Code agent.} We additionally evaluate Claude Code agent to compare our framework to a production agent harness. We evaluate it with Opus 4.8 model \citep{2026opuscard}, using the default effort setting (high). We set $max\_turns$ = 30 to match the budget of our BOED harness, and provide access to WebSearch, WebFetch, and CourtListener tools via an MCP connector. We use the same prompt as our non-agentic baselines to keep the comparison consistent. 

\textbf{Metrics.}
We treat the extraction of hallucinated segments from brief excerpts as a retrieval task. 
Let $G = \{g_1, \dots, g_m\}$ denote the set of ground-truth hallucinated segments in a brief excerpt and $P = \{p_1, \dots, p_n\}$ denote the set of segments predicted by the model.

Because predicted spans may not exactly match the annotated spans, we adopt a relaxed matching criterion. 
Two segments $p_1$ and $g_1$ are considered a match if either $p_1$ is a substring of $g_1$ or $g_1$ is a substring of $p_1$. 
This still counts predictions that slightly over- or under-span the annotated segment as correct.

Let $\text{match}(p, g)$ be an indicator function that equals 1 if predicted segment $p \in P$ matches a ground-truth segment $g \in G$ under this criterion and 0 otherwise. 
We compute the number of true positives as the number of predicted segments that match at least one ground-truth segment. We then compute:
(1) \textbf{Precision} $= {\text{TP}}/{|P|}$, the proportion of predicted segments that correspond to a ground-truth hallucinated segment. (2) \textbf{Recall} $= {\text{TP}}/{|G|}$, the proportion of ground-truth hallucinated segments that are correctly identified by the model. (3) \textbf{F1 score} $= \frac{2 \cdot \text{Precision} \cdot \text{Recall}}{\text{Precision} + \text{Recall}}$.

\subsection{Results}
\begin{table}[t]
\footnotesize
\centering
\begin{tabular}{lcccccc}
  \hline
  Model & \multicolumn{3}{c}{Agentic (max\_steps=30)} & \multicolumn{3}{c}{Non-Agentic} \\
   & Precision (\%) & Recall (\%) & F1 (\%) & Precision (\%) & Recall (\%) & F1 (\%) \\
  \hline
  Gemini 2.5 Flash & 16.9{\scriptsize $\pm$2.4} & 66.9{\scriptsize $\pm$5.1} & 27.0{\scriptsize $\pm$3.3} & 25.3{\scriptsize $\pm$4.2} &
  43.5{\scriptsize $\pm$5.9} & 32.0{\scriptsize $\pm$4.4} \\
  GPT-5 & 40.8{\scriptsize $\pm$4.3} & \textbf{84.4}{\scriptsize $\pm$4.2} & 55.0{\scriptsize $\pm$4.2} & 36.9{\scriptsize
  $\pm$4.5} & 58.2{\scriptsize $\pm$5.2} & 45.2{\scriptsize $\pm$4.2} \\
  GPT-OSS 120B & 21.1{\scriptsize $\pm$3.2} & 55.1{\scriptsize $\pm$5.8} & 30.5{\scriptsize $\pm$4.0} & 14.9{\scriptsize $\pm$3.0} & 30.7{\scriptsize
  $\pm$5.5} & 20.0{\scriptsize $\pm$3.7} \\
  Qwen3-8B & 12.0{\scriptsize $\pm$2.2} & 41.1{\scriptsize $\pm$6.0} & 18.6{\scriptsize $\pm$3.1} & 14.6{\scriptsize $\pm$2.8} & 35.1{\scriptsize
  $\pm$5.4} & 20.7{\scriptsize $\pm$3.5} \\
  Qwen3.6-27B & 28.9{\scriptsize $\pm$4.1} & 65.0{\scriptsize $\pm$5.1} & 40.0{\scriptsize $\pm$4.4} & 45.9{\scriptsize $\pm$7.8} & 31.5{\scriptsize $\pm$5.4} & 37.4{\scriptsize $\pm$5.7} \\
  \midrule
  Claude Code with Opus 4.8 $^\dagger$ & \textbf{76.1{}\scriptsize $\pm$5.6} & 62.8{\scriptsize $\pm$5.6} & \textbf{68.8}{\scriptsize $\pm$4.6} & -& - & - \\
  \hline
  \end{tabular}
\caption{\textbf{Agentic and non-agentic model performance.} All models have higher recall using the agentic framework, and GPT-5 achieves the best performance overall with BOED framework. $^\dagger$Claude Code uses its own native agent harness and achieves better precision and F1 score.}
\label{tab:hallucination_checker_results}
\end{table}                             
We show the main results in Table \ref{tab:hallucination_checker_results}, and detailed breakdown by hallucination type in Appendix Table \ref{tab:hallucination_checker_results_per_type}. The BOED agentic harness achieves higher recall than non-agentic experiments.

GPT-5 achieves the highest recall on all hallucination types. GPT-5 also has the second highest average number of steps per episode, averaging at 15.3 steps compared to the lowest, 7.5 steps, from Qwen3-8B. While evaluating agent trajectories, we find that GPT-5 performs the most thorough search, devoting 38.9\% of its actions to explore a previously retrieved opinion (action SEARCH\_\\LOCAL\_OPINION and READ\_DOCUMENT). It also more thoroughly verifies quotes and stated holdings, consistent with its higher recall on incorrect pincite, misquote, and content misrepresentation categories.

Claude Code agent achieves the highest precision and as a result, the highest F1 score, while its recall is lower than GPT-5, Gemini 2.5 Flash, and Qwen3.6-27B's. It uses tool calls more efficiently compared to GPT-5. Although its agent harness supports multiple tool calls per turn which allows it to have a higher tool call budget, it uses only an average of 15.4 tool calls per excerpt. 

Appendix Table \ref{tab:hallucination_checker_results_per_type} shows agents' recall performance on each hallucination type. All models, with the exception of Qwen3-8B, find most ( > 80\%) non-existing cases and case name mismatches, whereas no models can reliably detect wrong pincites. Page number information for many cases is only accessible through Westlaw and LexisNexis, which explains why models perform the worst on this category. Out of all the opinions retrieved by all agents, 19.9\% of them contain no usable text or lack pagination information. For content-based hallucinations (verbatim misquotes and content misrepresentation), we observe subantial variation in model performance: stronger models (GPT-5) can find more than 80\% of them, where weaker models only retrieve around 50\%.

\subsection{Error Analysis}
\label{sec:error_analysis}
To better understand the potential failure modes of agentic systems, we perform an error analysis on GPT-5 with the BOED harness, the best-performing system, examining all false negatives and false positives in the types non-existent citation, case name mismatch, and verbatim misquote, with the type inferred from the agent's final task beliefs. 

\textbf{False positives on citations absent from CourtListener.} The test set contains 132 correct distinct citations that return no lookup result on CourtListener. Any model that flags them produces a false positive because we do not alter these citations. Model behaviour on this subset varies substantially: Gemini 2.5 Flash flagged 65.9\% of these citations as hallucinated, while GPT-5 flagged only 25.0\% (Appendix Table \ref{tab:courtlistener_fp_rate}). Weaker models appear to treat absence from CourtListener as evidence of hallucination, while stronger models attempt to gather more information from alternative sources before reaching a verdict. Of the citations GPT-5 did not immediately flag, 40 were ultimately verified through alternative sources such as open CourtListener searches or web lookups, while 33 were left pending or unclear due to inconclusive evidence. We show examples of the GPT-5 agent leveraging alternative sources to verify citations in Appendix \ref{app:trajectory}.

\textbf{Exhausted verification budget, sometimes due to redundant actions.} 36.7\% of false negatives occur because the agent reached the maximum step limit of 30 before verifying the citation: for example, the missed citation was still unverified when the episode terminated. GPT-5 hit the step limit on 29.5\% of test episodes, which suggests that a higher step budget could further improve recall. At the same time, the GPT-5 agent does not use its budget efficiently in many episodes. We observe two main types of redundancies. First, when a citation lookup or content search reaches a dead end, the agent often re-issues the same query rather than concluding the citation as hallucinated. The duplicate citation lookup and duplicate opinion search appear in 39.7\% and 26.4\% of episodes respectively. Second, the agent sometimes re-queries an opinion they have already successfully searched with a slightly different query string. This sometimes destabilizes their belief after a prior search had returned a match. This is observed in 8.2\% of episodes (See Appendix Table \ref{tab:redundant_searches}). These redundant searches consume budget without making actual progress. In contrast, only 7.4\% of Claude Code’s calls are exact duplicates of earlier ones. However, 21.5\% of its CourtListener tool calls are invalid. This is due to agent guessing the tool calls based on its training data instead of calling tool search first to discover the available tools.

\textbf{Reliability failures.} Another significant failure mode is the agent's reliability issue (30.6\% of false negatives). A few examples include the following: (1) the agent returns output not conforming to the specified format, (2) the agent mistakenly returns early without verifying all identified citations, and (3) the agent fails to identify all content related to a citation. \patty{Update this}

Interestingly, agent verification also surfaced over 10 human-made citation typos present in the original pre-LLM briefs (See Appendix Table \ref{tab:gt_typo_corrections} for details).
\section{Discussion and Future Directions}
\label{sec:discussion}

\subsection{Information Access Limits Verification Performance}

Information access is a binding constraint on verification performance, independent of model capability. Although judicial opinions are public domain, they are not always freely accessible through a single centralized service. For example, the official federal system (PACER) charges per-page fees, and commercial providers routinely bill up to \$100 per query \citep{franklin2023lexiswestlaw}.While nonprofits like CourtListener have assembled large free databases of opinions, coverage is incomplete and contains no information about whether cited precedent is still good law. Niche cases generated by newer models are more likely to be absent from public repositories, especially if they contain Westlaw-specific identifiers. Additionally, public repositories sometimes lack official reporter pagination, which originates in commercial publisher formatting. This makes it more difficult to verify pincites using publicly available sources.

Automated verification tools inherit these same constraints around information accessibility. They directly explain our pincite results and account for a substantial share of false positives from weaker models. Weaker models would interpret absence of a cited case in CourtListener as direct evidence of a hallucination rather than a coverage gap. As shown in \S~\ref{sec:error_analysis}, the Gemini 2.5 Flash model falsely flagged 65.9\% of citations not available on CourtListener as hallucinated. Improving automated verification performance will thus require either broader public access to legal databases, or verification systems built on top of commercial platforms. Efforts to broaden free access to federal court records \citep{scales2024} are steps in the right direction, but the gap between what is publicly available and what is needed for reliable citation verification remains substantial. Policy efforts to improve equitable access to all case law and legal citations would reduce downstream burdens on the judiciary through improved automated verification methods like those we show here.

\subsection{Pro Se Litigants Need Better Support}

These information accessibility limitations fall disproportionately on pro se litigants, the group most likely to rely on LLMs for drafting and least likely to have access to commercial legal research databases.
The percentage of AI hallucinations in courts attributable to pro se litigants has been steadily rising every year. Courts already postulate that reasons for AI hallucinations in these cases are due to (1) a lack of awareness  \citep[See][]{ligeri2025amazon, alhamin2024star} and (2) a lack of access to effective verification practices  \citep[See][]{dukuray2024experion}. Courts have responded with sanctions, case dismissals, and, in some jurisdictions, mandatory AI disclosure requirements \citep{ropesgray2025tracker}. However, these responses are neither designed for nor directed at pro se litigants, and do not foster the potential of AI to increase access to justice for this group.

First, we need effective interventions that educate pro se litigants on how to use AI tools responsibly and equip them with  best practices and tools for verifying AI-generated citations. Courts already provide resources for pro se litigants.
These should include risks of using AI for pro se litigants, and include detailed AI literacy and citation verification.
Second, expanding public, freely-available access to legal data (including all case law) would improve both human and automated citation verification performance for under-resourced filers. For example, Canada's 
legal information infrastructure is organized around a single free public database unlike the United States. CanLII, a non-profit maintained by the Federation of Law Societies of Canada, provides free access to court judgments from all Canadian courts and all jurisdictions, with generally complete coverage for cases decided after 2001 \citep{canlii2025about}. This structural 
difference is reflected in how courts reason about pro se verification: one Canadian court noted that verifying whether AI-cited cases exist requires only ``a simple search on 
CanLII,'' a baseline assumption that US courts cannot reasonably make. \cite{ok2025southern}.
Third, automated verification tools could ease verification burdens of courts, with sufficient access to data, but also assist non-lawyers directly to verify citations in documents drafted with AI assistance. Our benchmark provides a step towards developing and auditing such tools.

If courts simply continue to raise sanctions, while not addressing structural limitations around citation verification, AI will ultimately be a false promise for pro se litigation. Facing significant sanctions for potential AI hallucinations and no reliable way to verify citations, pro se litigants may eventually abandon AI altogether. This is in contrast to professional parties who can get access to AI for legal research, and the required tools and practices to ensure responsible usage. Moreover, if access to legal databases becomes a prerequisite for responsible AI use, hiring a professional may ultimately be cheaper and less risky than litigating pro se.

\subsection{Future Directions}
Despite these information access constraints, we find that agentic systems can already be useful. The agent surfaced over 10 citation typos in pre-LLM briefs that human authors missed. As the volume of AI-generated filings grows, citation verification becomes a more central task. Law clerks, who assist judges by reviewing submitted filings and verifying legal arguments, may take on an expanded role as a result \citep{katrak2025criticalpeople}. Our findings suggest that agentic systems can support this work by flagging errors in draft briefs before filing or assisting law clerks in screening submitted documents. However, reliability issues and long-context failures persist and require more research.

The primary aim of this benchmark is to support the development of more reliable and performant citation verification methods. It can further be used to audit existing citation verification tools to inform procurement processes of public agencies. However, the semi-synthetic nature of our benchmark means that results may not fully generalize to the distribution of hallucinations in real filings. We therefore encourage future work to complement our dataset with naturally occurring examples. Our benchmark should primarily be treated as a controlled lower bound on verification difficulty.

There are several future directions. In this work, we evaluate the models on short excerpts, but real legal briefs are far longer. The longest brief we collected contains 193 pages, more than 30{,}000 words, and over 1{,}000 citations. Agents' problem with using compute resource efficiently and behaving reliably are both exacerbated in citation-dense briefs \citep{rabanser2026scienceaiagentreliability}. Hence, future work on legal hallucination detection should focus on alleviating these two issues.

\section{Related Work}
\label{sec:related}

As LLMs are increasingly applied to the legal domain, numerous works have developed benchmarks and evaluations for hallucinations in legal tasks. These include closed-domain legal summarization~\citep{deroy2023readypretrainedabstractivemodels,2021improving}, legal concept explanation~\citep{savelka2023explaininglegalconceptsaugmented}, and open-domain legal question answering~\citep{Dahl_2024,hu-etal-2025-fine, fan2026halluhardhardmultiturnhallucination}. Prior work has also proposed taxonomies of legal hallucinations: \citet{Dahl_2024} introduces a taxonomy of general legal hallucinations, \cite{magesh2025hallucination} expands it to apply to RAG systems, and \cite{hou-etal-2024-gaps} characterize ``gaps'' between human-written and machine-generated legal analysis, including hallucinations. These benchmarks and taxonomies primarily focus on the correctness of generated legal text. In contrast, our work targets hallucinations in case law citations embedded within legal documents, which is the most common hallucination observed in courts. The closest work to ours evaluates whether LLMs can generate correctly formatted Bluebook citations \citep{dahl2025byebyebluebookautomatinglegal}. This work does not focus on the correctness of citation format, which is a close-domain task, but rather if the citation contains the correct information. 

An alternative to reducing hallucinations through model-level improvements is to apply post-hoc safeguards that verify or correct model outputs after generation. 
Prior work has explored a range of post-hoc hallucination detection and correction methods \citep{chakraborty2025hallucination}. These approaches can be grouped into methods that require access to the original generation process, such as generating multiple responses \citep{manakul-etal-2023-selfcheckgpt, mundler2024selfcontradictoryhallucinationslargelanguage, yehuda-etal-2024-interrogatellm} or using specialized prompting strategies \citep{yu-etal-2024-reeval}, and methods that operate independently of the generator. The latter category includes training external discriminators to detect hallucinations~\citep{2023reid} and systems that query external sources to verify or correct generated summaries~\citep{chen2023purrefficientlyeditinglanguage}. Our work aligns with this second class of approaches by treating hallucination detection as a downstream verification problem.

\section{Conclusion}
Legal citation hallucinations are not a temporary artifact of early LLMs. Across eight ChatGPT generations, we find that legal citation hallucination rates are not consistently decreasing, and that the verification burden on courts is growing along two compounding dimensions: more filings and more citations per filing. These trends will not self-correct as models improve. We thus argue that attention must shift to verification.

We introduce a taxonomy of legal citation hallucinations grounded in real court filings, a dataset of 1,300 brief excerpts with injected hallucinations, and we evaluate an agentic verification system that substantially improves recall over non-agentic baselines. Our results show that there is room for improvement: the best model achieves 68.8\% F1, and content misrepresentation — the hallucination type most likely to distort legal outcomes — remains the hardest to detect across all systems. Structural barriers, particularly the incompleteness of official pagination in publicly available legal repositories, further limit what automated verification can currently achieve.

Responsible deployment of AI in legal contexts requires treating citation verification as a first-class problem. Our taxonomy, dataset, and evaluation framework are a step toward that goal, providing a foundation for building, benchmarking, and auditing the citation checking tools that courts and litigants increasingly need.

\newpage
\bibliographystyle{plainnat}
\bibliography{references}

\newpage
\beginsupplement
\section{Dataset Construction}
We describe the dataset curation process in more detail in this section. 

\subsection{Bluebook Legal Citation Format}
\label{app:bluebook}
Verifying a legal citation is not a single check but a structured set of checks over each component that the citation encodes. Legal citations in U.S. court filings follow the Bluebook, a standardized citation system used across courts and legal practice \citep{bluebook2025}.
A standard case citation consists of five components: (1) a \emph{case name}, identifying the parties, (2) a \emph{reporter citation}, specifying the volume, reporter, and first page of the decision, (3) an optional \emph{pincite}, indicating the specific page(s) supporting a proposition or quotation, (4) a \emph{parenthetical}, providing the court and year of decision, and (5) other parenthetical information, and subsequent history of the case, if any. For example, in a citation of the form \emph{Aves v. Shah}, 997 F.2d 762, 767 (10th Cir. 1993), \emph{Aves v. Shah} is the case name, 997 F.2d 762 is the reporter citation, 767 is the pincite of a particular page, 10th Cir. is the court, and 1993 is the date of decision. All components should be consistent to form a correct citation.

\subsection{Hallucination Injection Mechanism}
\label{app:injection_mechanism}
In this section we detail the hallucination injection mechanism. We inject hallucination into 80\% of the briefs. We decide to use 80\% instead of 50\% because a brief that contains hallucination would still result in many segments that are hallucination free. To ensure that there are enough segments with hallucinations to sample from, we choose a higher injection rate. Then, for each brief that we inject hallucinations into, we randomly select a number between 0 and the number of citations in the document as the global budget for the number of hallucinations. We randomly sample citations for hallucination, and use weighted random sampling to select the hallucination type to inject. We use weighted sampling to choose hallucination type to ensure balanced distribution across all types because not all case citations are eligible for all types of hallucination. For example, only citations that are associated with quotes are eligible for misquotes, while all citations are eligible for non-existent citation. Additionally, we apply non-existent citations and case name mismatches globally across all instances of the same case citation. We apply these globally to ensure that the verification model cannot infer hallucination based on correct citations of the same case that appear in other parts of the brief. For example, one instance of the citation "\textit{Cinel v. Connick}, 15 F.3d 1338" is altered to "\textit{Boone v. Vinson}, 15 F.3d 1338", but if the correct citation "\textit{Cinel v. Connick}, 15 F.3d 1338" appears in another sentence, the model might be able to infer that the citation contains some inaccuracies solely based on the inconsistency. We expand on the specific injection mechanism for the five hallucination types our taxonomy defines below.

\paragraph{Non-existent Citation.} All citations are eligible for non-existent citation hallucination injection. We apply this type of hallucination globally. We use a list of non-existent court reporters and randomly generate numbers between 1 and 999 for the volume and first page number to generate non-existent citations. Using non-existent court reporters ensures that the generated citation is completely false. 

\paragraph{Mismatched Case Name.} Citations that appear together with their corresponding case names are eligible for mismatched case name hallucination injection. We apply this type of hallucination globally. We use \texttt{Qwen3-32B} to extract case names (model prompt in Appendix \ref{app:dataset_prompt}). We use two different methods to generate mismatched case name: a) altering case name and b) altering reporter citation. We first create a dataset of real cases that we can sample from for these alterations. We use the first 5{,}000 rows in \citet{harvard_lil_cold_cases} Collaborative Open Legal Data (COLD) Cases dataset and remove entries with empty \textit{case\_name column} and \textit{case\_name\_short} column. We randomly sample from the real case dataset entries to either replace the case name or the reporter citation. We also ensure that the sampled case name is distinct from the original correct case name (<0.9 similarity using fuzzy matching) and the sampled reporter citation is distinct from all reporter citations of the original correct citation (list of reporter citations obtained from CourtListener lookup). 

\paragraph{Incorrect Pincite.} Citations that contain pincites are eligible for incorrect pincite hallucination injection. CourtListener citation lookup API extracts the reporter citations but does not extract the pincites. We implement a regular expression matching function to extract pincites from various case citation formats. After extracting the original correct pincites, we alter the first pincite to be a different number between the first page and the original pincite. We use this method to ensure that the new false pincite still belongs to the same opinion, to not overlap with other hallucination types.

\paragraph{Verbatim Misquote.} We use \texttt{Qwen3-32B} to extract and alter quotes (model prompts in Appendix \ref{app:dataset_prompt}). We ensure that the altered quotes are different from the original correct quotes, and that the lengths are similar (between 0.7 and 1.3 or the original). We also ensure that both the extracted quotes and the altered quotes are enclosed in quotation marks.

\paragraph{Content Misrepresentation.} We use \texttt{Qwen3-32B} to extract and alter quotes (model prompts in Appendix \ref{app:dataset_prompt}). We ensure that the altered holdings are different from the original correct holdings. Due to the expertise required in annotating this, we had a second senior expert author review a subset of 40 content misrepresentation annotations and we agree on 95\% of these with the remainder being borderline cases.

\subsection{Legal Hallucination Dataset}
\label{app:dataset}
We collect a dataset of legal briefs that were filed in courts. We use the CourtListener search API to find federal appellate briefs that were filed between 2012-01-01 and 2021-12-31. We downloaded 323 documents in total in PDF format. We then convert all PDFs to plain text using olmOCR \citep{olmocrbench}.

\begin{figure}[h]
    \centering
    \includegraphics[width=0.7\linewidth]{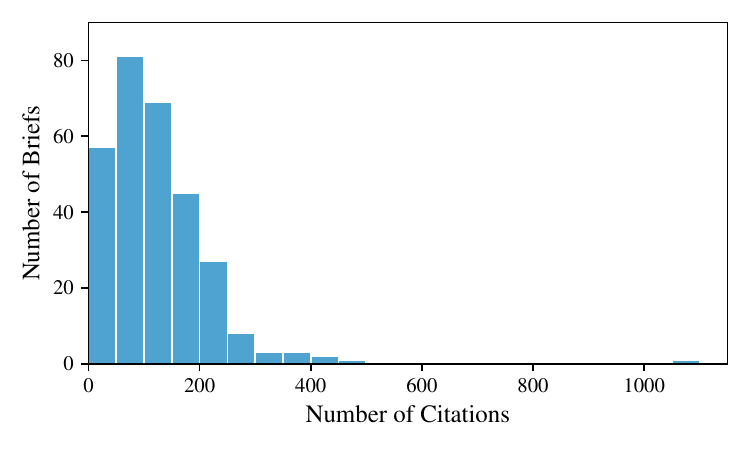}
    \caption{\textbf{Number of citations per brief.} Most briefs have 0-200 citations.}
    \label{fig:citation_per_brief}
\end{figure}

To construct the hallucination dataset, we proceed in several steps. First, we use the CourtListener citation lookup API to extract all case citations from the collected briefs, yielding a total of 27{,}949 distinct legal citations. 2{,}854 citations of these are not found on CourtListener. Most of these citations are Westlaw or Lexis citations. We do not alter any citations that are not found on CourtListener because these citations are already challenging to verify given the lack of access. Figure \ref{fig:citation_per_brief} shows the number of citations per brief distribution. Most briefs have between 0 to 200 citations. We discard briefs with over 200 citations, leaving 279 briefs. After injecting hallucinations and breaking briefs into paragraphs, we subsample again to form the final dataset that contains 1{,}000 excerpts. The excerpts in the final dataset come from 245 briefs.

For each citation, we extract a local context from the brief. Next, we use the \texttt{Qwen3-32B} \citep{yang2025qwen3technicalreport} models to identify the associated case name and, when present, the corresponding quotation as well as the holding each citation is used to support based on the surrounding context. To decrease the risk of hallucination, we use LLMs solely to identify and retrieve structured citation components from the original briefs. We also ensure that all the retrieved case names, quotations, and holdings appear verbatim in the original text. 

Finally, we inject hallucinated citations by systematically modifying one or more components of the original citation according to the taxonomy in Section \ref{sec:taxonomy}. We detail the injection mechanism in Appendix \ref{app:injection_mechanism}.

\begin{figure}
    \centering
    \includegraphics[width=0.95\linewidth]{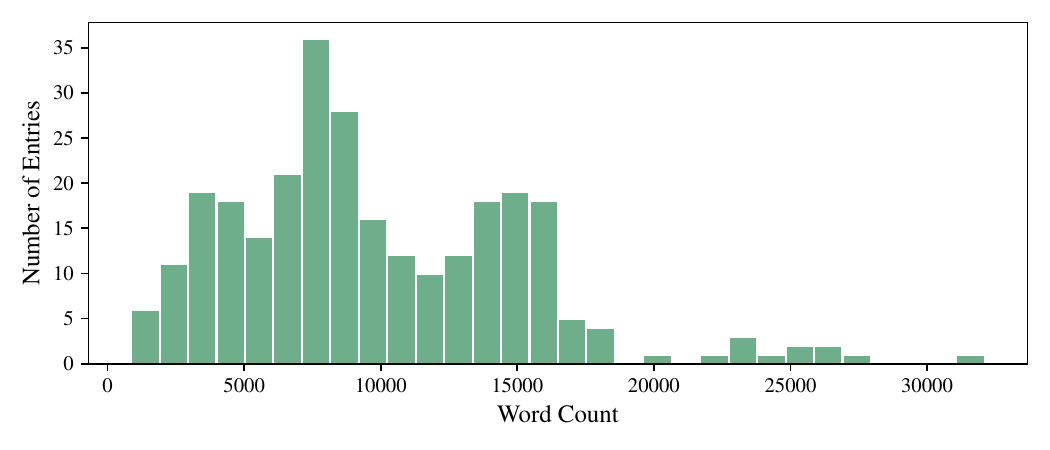}
    \caption{\textbf{Word count per brief.} The variance in word count of briefs is very high.}
    \label{fig:brief_word_count}
\end{figure}

Then, we parse full briefs into segments. We use segments instead of full briefs as unit for verification to standardize the process. The briefs have a wide range of word counts (See Figure \ref{fig:brief_word_count}).
We apply LLM-based semantic segmentation \citep{smith2024evaluating}: We first fine-tune a RoBERTa model to do legal sentence segmentation \citep{sanchez-2019-sentence}. We then use this model to segment briefs into sentences. Afterwards, we use a Llama3.3-70B model to group sentences into semantically coherent paragraphs, resulting in 5{,}648 segments. We subsample 1,000 from all resulting segments. When sampling, we exclude all segments that contain Table of Content because they only contain lists of case citations without any context, making them less challenging. We also exclude contact information of attorneys and only sample segments that contain at least one citation.

We only use the central holding task portion of the Legal Hallucinations dataset from \cite{Dahl_2024} because it is the closest to our task. However, it is a QA dataset, so we need to re-format it to be aligned with our dataset input. The original dataset is constructed by prompting LLMs to output the primary holding of a case given its reporter citation and year. It contains 1494 entries based on 300 unique cases and has outputs from 4 LLMs. Each entry has a binary \textit{hallucination} label, a \textit{correctness\_score} ranging from 0-100, additionally with -99 indicating invalid response, and two LLM responses \textit{llm\_output\_1} and \textit{llm\_output\_2}.

Each case has multiple entries corresponding to responses from the different LLMs. We subsample from the whole dataset so that each case has one entry, resulting in 300 entries. When subsampling, we prioritize sampling entries with no hallucination to result in a more balanced final dataset. There are 115 non-hallucinated responses and 185 hallucinated responses in the subsampled dataset according to the original \textit{hallucination} label. We transform the dataset format to align with our dataset's. We construct each entry:

\begin{quote}
\textbf{Hallucinations:} [$<$\textit{llm\_output\_1}$>$] if \textit{hallucination} is True, otherwise [].

\par
\textbf{Text:} ``$<$\textit{llm\_output\_1}$>$. See $<$\textit{case name}$>$, $<$\textit{citation}$>$ ($<$\textit{year}$>$).''
\end{quote}

\textit{llm\_output\_1}, \textit{citation}, and \textit{year} are fields in the Legal Hallucinations dataset, and \textit{case name} is obtained from a CourtListener lookup of the citation.

However, the \textit{hallucination} labels are not based on whether the LLM outputs are correct but whether the two outputs are consistent with each other. This leads to a lower bound measure of hallucination rate, but for our dataset we need accurate hallucination labels, so we manually verify all labels. 

We verify all 300 entries using Westlaw. We leave the text as is if it is consistent with its \textit{hallucination} label. If there is inconsistency in the stated case year and the Westlaw opinion year, we use the year from Westlaw. 

Combining the two parts of the dataset, there are 1{,}300 entries in total. 

\begin{figure}
    \centering
    \includegraphics[width=\linewidth]{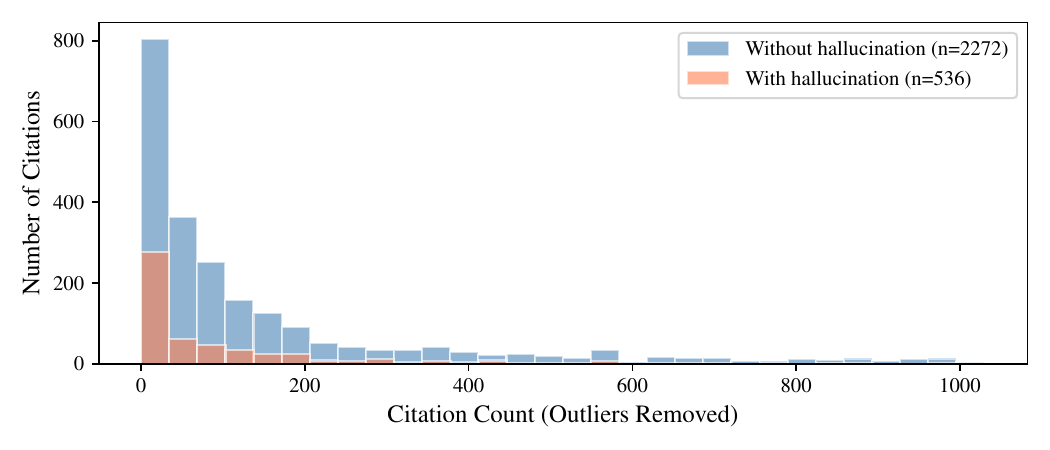}
    \caption{\textbf{Number of citations cases have.} The hallucinated (red) and non-hallucinated (blue) cases have the same citation distributions. We remove the outliers when plotting because a few cases are cited many times that they would skew the visualization.}
    \label{fig:citation_of_cases}
\end{figure}

Figure \ref{fig:citation_of_cases} shows the distribution of the number of times each case in the dataset has been cited. We see that both hallucinated and non-hallucinated cases follow the same citation distribution. The model thus cannot solely rely on knowledge of landmark cases to check for hallucinations.

\paragraph{Example datapoint.}
Each entry in the dataset consists of a short excerpt from a legal brief and a list of ground truth hallucinated segments with the hallucination type label. The task is to extract all hallucinated segments from the excerpt. Below is an illustrative example involving incorrect pincite and content misrepresentation:

\begin{quote}
\small
\textbf{Hallucinations:} \{"370 F.3d 1223, 1224": "incorrect\_pincite", "US Parole Commission members are 1983 persons when they act pursuant to the Eighth Amendment’s prohibition on cruel and unusual punishment.": "Content misrepresentation"\} \\
\textbf{Text:}  
``[...] but the Commissioners exercise District powers when they handle D.C. parolees.  See, e.g., Fletcher v. Dist. of Columbia, 370 F.3d 1223, 1224 judgment vacated on reh'g on other grnds, 391 F.3d 250 (D.C. Cir. 2004) US Parole Commission members are 1983 persons when they act pursuant to the Eighth Amendment’s prohibition on cruel and unusual punishment. [...]''
\end{quote}

In this example, 370 F.3d 1223, 1224 is hallucinated because the pincite is incorrect, and the holding is misrepresented because it should be "pursuant to D.C. Revitalization Act" instead of "pursuant to the Eighth Amendment’s prohibition on cruel and unusual punishment". 

\subsection{Dataset Label Update}
We create the ground truth hallucination segments of each dataset entry under the assumption that the original briefs do not contain any mistakes that are similar to model hallucinations. We select briefs from 13 U.S. Courts of Appeals to ensure quality. However, humans make mistakes as well. We examined all false positive outputs from GPT-5 agent of the types non-existent citation, case name mismatch, and misquote (hallucination types are inferred from the final task beliefs). We found some typos in the original briefs, as well as some minor stylistic choices that are ambiguous in terms of whether they should be counted as mistakes or not. We update the ground truth hallucination labels to include the clear typos (See Table \ref{tab:gt_typo_corrections}). We create another category of optional ground-truth labels to include stylistic choices (e.g., "fact checking" vs. "factchecking") and other instances where we cannot locate the case on Westlaw but are also not confident that the case does not exist (see Table \ref{tab:optional_gt}). An optional ground truth segment means that a prediction matching these spans counts as a true positive, but failing to predict them does not count as a false negative. 

\begin{table}[t]
\centering
\scriptsize
\begin{tabular}{lll}
\hline
\textbf{File} & \textbf{Original} & \textbf{Correct} \\
\hline
  \texttt{Abbey v United States} & Gieg v. DRR, Inc. & Gieg v. DDR, Inc. \\
  \texttt{Abbey v United States} & 216 F.3d 659 & 276 F.3d 659 \\
  \texttt{Crystallex Int'l v Petroleos} & Shenago, Inc. & Shenango, Inc. \\
  \texttt{Delaware Riverkeeper v FERC} & egregious or arbitrary & {[egregious]} or {[arbitrary]} \\
  \texttt{Gametek v Zynga} & `so theoretical and broad' & `so abstract and sweeping' \\
  \texttt{Gametek v Zynga} & Benson, 409 U.S. at 67 & Benson, 409 U.S. at 68 \\
  \texttt{In re Blaine Keith} & 587 U.S. 460 & 567 U.S. 460 \\
  \texttt{In re Blaine Keith} & Fuentes v. Shervin & Fuentes v. Shevin \\
  \texttt{Intercollegiate Broadcasting v CRB} & vesting authority & vesting of authority \\
  \texttt{Jones v Citimortgage} & Bonano v. Thomas & Bonanno v. Thomas \\
  \texttt{Loveridge v Hall} & ``jurisdiction over all'' & ``jurisdiction of all'' \\
  \texttt{Nozzi v HACLA} & avoiding an abrupt change & an abrupt and unexpected change \\
  \texttt{United States v Esquenazi} & Id. at 815-16 & Id. at 809 \\
  \texttt{Westberg v FDIC} & 624 F.3d at 1144 & 642 F.3d at 1144 \\
  \texttt{Witham v New York State} & Roadway Express v. Pipe & Roadway Express v. Piper \\
  \texttt{Witham v New York State} & Sims v. Aherns & Sims v. Ahrens \\
  \texttt{Zzyym v Pompeo} & Plyer v. Doe & Plyler v. Doe \\
\hline
\end{tabular}
\caption{\textbf{Typographic errors found in original briefs} We found these minor errors in the original briefs when examining the false positives in agent outputs. We update the ground truth labels to include them.}
\label{tab:gt_typo_corrections}
\end{table} 

\begingroup
\scriptsize
\begin{longtable}[t]{>{\raggedright}p{3.5cm}l p{4.5cm}}
\centering
\textbf{File} & \textbf{Type} & \textbf{Optional Span} \\
\hline
\texttt{Allina Health v Sebelius} & Case name mismatch & Heartland Regional Med.Ctr. v. Sebelius \\
\texttt{Kimberlin v Frey} & Verbatim misquote & a loan servicer will become a debt collector under 1692a(6)(F)(iii) if the debt was in default or treated as such when it was acquired. \\
\texttt{Clear with Computers v Altec} & Verbatim misquote & meaningful limitations \\
\texttt{Estate of Bonilla v City of York} & Verbatim misquote & raise an inference \\
\texttt{Evolutionary Intelligence v Sprint} & Content misrepresentation & involves a specific system for modifying data that has equally concrete and valuable effects in [its field]. \\
\texttt{Evolutionary Intelligence v Sprint} & Incorrect pincite & Diehr, 101 S.Ct. at 1048 \\
\texttt{Fortres Grand v Warner Bros.} & Verbatim misquote & unfair competition category \\
\texttt{Carrera v Bayer Corp} & Non-existent citation & 135 S. Ct. 940 (2013) \\
\texttt{Carrera v Bayer Corp} & Verbatim misquote & best practicable notice \\
\texttt{Gonzalez v New Life Ventures} & Verbatim misquote & It is obvious that the subject matter described in 101 is expansive. As the Supreme Court has observed, the subject-matter provisions of the patent law have been cast in broad terms to fulfill the constitutional and statutory goal of promoting the Progress of Science and the useful Arts. \\
\texttt{Hockeyline v Stats LLC} & Verbatim misquote & this subsidiary fact finding must be reviewed for clear error on appeal. \\
\texttt{Nemphos v Nestle Waters N. Am.} & Verbatim misquote & ``Pure Water Perfect Taste'' \\
\texttt{Rentmeester v Nike} & Verbatim misquote & a court must filter out and disregard the non-protectable elements in making its substantial similarity determination. \\
\texttt{Jones v Citimortgage} & Verbatim misquote & a weighing of evidence on both sides, \\
\texttt{Jones v Citimortgage} & Verbatim misquote & mechanically determined through relatively rigid legal rules. \\
\texttt{Westberg v FDIC} & Verbatim misquote & As a practical matter of statutory construction, \ldots\ we proceed on the assumption that Congress intended the `claims' barred by 1821(d)(13)(D) to parallel those contemplated under FIRREA's administrative claims process laid out in the greater part of 1821(d). \\
\texttt{Belleau v Wall} & Non-existent citation & Ellis v. State, 353 S.E.2 19 (Ga. 1987) \\
\texttt{Par Pharmaceutical v TWI} & Verbatim misquote & {[u]}se of prophetic examples does not automatically make a patent non-enabling. \\
\texttt{Smith v Aegon Pension Plan} & Verbatim misquote & are generally free under ERISA, for any reason, at any time, to adopt, modify, or terminate welfare plans \\
\texttt{United States v Velasquez} & Content misrepresentation & The prosecutor's comment during closing argument was more prejudicial. It was a direct comment on defendant's failure to testify, and it occurred only one day before the jury began deliberations. \\
\texttt{United States v Flanders} & Verbatim misquote & In general, a sentence within the limits imposed by statute is neither excessive nor cruel and unusual. \\
\texttt{W. Reserve Life v DK LLC} & Non-existent citation & 715 F. Supp. 270 (D.R.I. 2010) \\
\hline
\caption{\textbf{Optional ground-truth segments.} A prediction matching these spans counts as a true positive, but failing to predict them does not count as a false negative.}
\label{tab:optional_gt}
\end{longtable}
\endgroup

\subsection{Controlled Experiment LLM-Generated Legal Documents}
We generate prompts in this experiment by collecting legal filings with AI-generated hallucinations and reverse-engineering the prompts that were used to generate them. We show example prompts in Table \ref{tab:legal_doc_prompt}.

\begin{table}[t]
\centering
\footnotesize
\begin{tabular}{p{0.25\linewidth} p{0.7\linewidth}}
\toprule
 & Prompt Text \\
\midrule
Prompt example 1 & Generate an Objection to Report and Recommendation in the Denial of In Forma Pauperis Application.  Cite relevant cases to support the objection. The following facts are given: \newline
The plaintiff, Bonnie Nicole Luther, is objecting to a Magistrate Judge's recommendation to deny her application to proceed in forma pauperis (without prepayment of fees or costs). She argues that she cannot afford the filing fees and should be granted IFP status.
Her reasons for objecting include: \newline
1. Lack of Financial Capacity: She claims to have no personal income or public benefits, and although she has a joint bank account with her husband, the funds are exclusively for essential household expenses for their family of five, and she lacks control over discretionary finances. She contends that mere joint account access does not equate to disposable income. \newline
2. Joint Property Status: While she is a joint owner of her home, she states she receives no financial benefit from it, has no accessible equity, and it is a primary residence, not a liquid asset. \newline
3. Good Faith Submission: She asserts that she completed her IFP applications in good faith, responding based on her actual knowledge and access.\newline
4. Importance of Legal Claims: The underlying lawsuit involves serious allegations of civil rights and constitutional violations against state actors under federal law, and denying IFP status would prevent her from pursuing these claims due to poverty. She emphasizes that the in forma pauperis statute is meant to ensure equal access to the judicial system for indigent persons, especially in cases involving governmental abuse or constitutional harm. \\
\hline
Prompt example 2 & Generate an Objection to the Magistrate Judge’s Report and Recommendation Denying In Forma Pauperis Application.  Cite relevant cases to support the objection. The following facts are given:
The plaintiff, Bonnie Nicole Luther, is objecting to a Magistrate Judge's recommendation to deny her application to proceed in forma pauperis. She asserts the denial was incorrect and offers the following reasons: \newline
1. Irregular and Insufficient Income: She works part-time as a substitute teacher but earns well below the federal poverty line. Her hours are unpredictable, and during several months she earns no income at all. She argues this erratic income should not disqualify her from IFP status, especially as she has no savings and is behind on several utility and medical bills.\newline
2. Debt and Financial Obligations: Bonnie carries significant unsecured debt, including past-due credit card balances and a defaulted student loan. Additionally, she is the primary caregiver for an elderly parent with no income of their own, which imposes significant caregiving and financial burdens.\newline
3. Minimal Liquid Assets: While she does have a small individual checking account, the balance rarely exceeds \$50 and is depleted monthly by basic living expenses. She owns no property, has no investments, and does not own a car.\newline
4. Good Faith and Transparency: She emphasizes that she disclosed all requested financial information and attachments with full transparency, and no part of her application was intended to mislead or omit material facts.\newline
5. Meritorious and Time-Sensitive Legal Claims: Her lawsuit raises important claims under the Americans with Disabilities Act and Section 1983 for denial of reasonable accommodations and violations of her First and Fourteenth Amendment rights. She contends that denying her access to court based solely on poverty is contrary to the purpose of § 1915 and would irreparably harm her ability to seek redress for ongoing constitutional violations. \\
\bottomrule
\end{tabular}
\caption{Example prompts used to generate legal documents for LLM legal citation hallucination rate estimation.}
\label{tab:legal_doc_prompt}
\end{table}

\section{Additional Agentic Framework Details}

\subsection{Agent Actions}
\begin{table}[ht]
\footnotesize
\centering
\begin{tabular}{llp{6.5cm}}
\toprule
\textbf{Type} & \textbf{Action} & \textbf{Description} \\
\midrule
\multirow{2}{*}{\textit{Lookup}}
  & \texttt{COURTLISTENER\_CITATION\_LOOKUP} & Calls the CourtListener lookup API endpoint given a string of text containing a case citation. \\[3pt]
  & \texttt{OPEN\_COURTLISTENER\_SEARCH}      & Performs an open search on CourtListener using keywords. \\[3pt]
  & \texttt{OPEN\_WEB\_SEARCH}                & Searches the web with SerpAPI for information. \\[3pt]
\midrule
\multirow{2}{*}{\textit{Search}}
  & \texttt{ACCESS\_COURTLISTENER\_OPINION}   & Calls the CourtListener opinion API endpoint to access an opinion. \\[3pt]
  & \texttt{SEARCH\_LOCAL\_OPINION}           & Searches text in the local opinion. \\
\midrule
\multirow{2}{*}{\textit{Reading}}
  & \texttt{READ\_DOCUMENT}                   & Reads selected lines from the chosen opinion. \\[3pt]
  & \texttt{EDIT\_SCRATCHPAD}                 & Takes notes when reading documents. \\
\midrule
\multirow{2}{*}{\textit{Reasoning}}
  & \texttt{THINK}                            & Performs internal reasoning. \\[3pt]
\bottomrule
\end{tabular}
\caption{\textbf{Agent action types and their descriptions.} The agent has access to eight different actions grouped into four types.}
\label{tab:actions}
\end{table}

\subsection{Model Specification}

 \begin{table}[h]                                                                      
  \centering                                                                                                      
  \label{tab:model-configs}                                                       
  \begin{tabular}{llllccc}                                        
  \toprule                                                                              
  \textbf{Model} & \textbf{Access} & \textbf{Params} &
  \textbf{Temp.} & \textbf{Max Tokens} & \textbf{GPUs}
   & \textbf{Quant.} \\                           
  \midrule
  Gemini 2.5 Flash   & API (Gemini)   & ---   & 0.8 &       
  8,192 & -- & --\\                                                              
  GPT-5              & API (OpenAI)   & ---   & 0.8 &
  10{,}000 & -- & -- \\                                                              
  GPT-OSS-120B       & Local (vLLM)   & 120B  & 0.8 &
  8,192 & 2 & MXFP4 \\                                                              
  Qwen3.5-27B (FP8)  & Local (vLLM)   & 27B   & 0.8 &
  8,192 & 1 & FP8  \\                                                              
  Qwen3-8B           & Local (vLLM)   & 8B    & 0.8 &
  8,192  & 1 & MXFP4 \\                                                              
  \bottomrule                                                     
  \end{tabular}  
  \caption{\textbf{Model configurations used in experiments.}}  
  \end{table} 
\section{Additional Experimental Results}
\subsection{Model Performance by Hallucination Type}
Table \ref{tab:hallucination_checker_results_per_type} shows the agents' performance on different hallucination types. Most models achieve high recall on non-existent citations and case name mismatches

\begin{table}[t]
\scriptsize
\centering
\begin{tabular}{lcccccc}
  \hline
  Hallucination Type & Gemini 2.5 Flash & GPT-5 & GPT-OSS 120B & Qwen3-8B & Qwen3.6-27B\ & \makecell{Claude Code \\ with Opus 4.8
  $^\dagger$} \\
  \hline
  Non-Existent Citation & \textbf{100.0}{\scriptsize $\pm$0.0} & \textbf{100.0}{\scriptsize $\pm$0.0} & 83.9{\scriptsize $\pm$14.7} & 58.1{\scriptsize $\pm$18.5} & \textbf{100.0}{\scriptsize $\pm$0.0} & 93.5{\scriptsize $\pm$8.6} \\
  Case Name Mismatch & \textbf{100.0}{\scriptsize $\pm$0.0} & \textbf{100.0}{\scriptsize $\pm$0.0} & 84.2{\scriptsize $\pm$10.0} & 49.1{\scriptsize $\pm$15.2} & 96.5{\scriptsize $\pm$4.7} & 83.8{\scriptsize $\pm$9.9} \\
  Incorrect Pincite & 18.9{\scriptsize $\pm$10.2} & \textbf{52.8}{\scriptsize $\pm$16.1} & 26.4{\scriptsize $\pm$12.5} & 18.9{\scriptsize $\pm$9.8} & 50.9{\scriptsize $\pm$13.7} & 3.6{\scriptsize $\pm$14.0} \\
  Verbatim Misquote & 78.6{\scriptsize $\pm$12.8} & \textbf{95.2}{\scriptsize $\pm$6.1} & 42.9{\scriptsize $\pm$15.8} & 31.0{\scriptsize $\pm$14.3} & 64.3{\scriptsize $\pm$14.8} & 71.7{\scriptsize $\pm$7.9} \\
  Content Misrepresentation & 60.3{\scriptsize $\pm$8.6} & \textbf{83.2}{\scriptsize $\pm$6.4} & 51.1{\scriptsize $\pm$8.4} & 45.8{\scriptsize $\pm$9.0} & 48.9{\scriptsize $\pm$8.7} & 66.4{\scriptsize $\pm$4.5} \\
  \hline
  \end{tabular}
\caption{\textbf{Agentic model recall by hallucination types.} All models perform the worst on content misrepresentation and incorrect pincite tasks. We only show the recall in this table because the models do not lable the hallucination type when outputting hallucinated segments, thus making per type precision inaccurate.}
\label{tab:hallucination_checker_results_per_type}
\end{table}

\subsection{Agent Tool Call Analysis}
Table \ref{tab:action_type_frequency} shows the distribution of action types taken by each model during evaluation. Across all models, the three most frequently used actions are COURTLISTENER\_CITATION\_LOOKUP, ACCESS\_COURTLISTENER\_OPINION, and SEARCH\_LOCAL\_OPINION, reflecting that legal database access is the core bottleneck in citation verification. The primary difference between stronger and weaker models is not the number of lookups performed, but the actions taken after a successful lookup. GPT-5 devotes 37.3\% of its actions to SEARCH\_LOCAL\_OPINION, the highest of any model, while Qwen3-8B, the weakest performer, spends only 14.9\% on this action and instead concentrates over half its steps on direct citation lookups. This pattern suggests that after locating a citation, stronger models continue into the opinion text to verify quotes and holdings thoroughly. This is consistent with GPT-5's stronger recall on incorrect pincite, misquote, and content misrepresentation categories, all of which require reading the underlying opinion rather than merely confirming a citation exists. GPT-5 also uses OPEN\_WEB\_SEARCH more than any other model (8.3\% of actions), likely as a fallback strategy for citations absent from CourtListener. This behavior is examined further in the error analysis below.

\begin{table}[t]
\scriptsize\centering
\begin{tabular}{lrrrrr}
\hline
Action Type & Gemini 2.5 Flash & GPT-5 & GPT-OSS-120B & Qwen3-8B & Qwen3.6-27B \\
\hline
ACCESS\_COURTLISTENER\_OPINION & 24.7\% & 19.5\% & 17.6\% & 16.8\% & 24.8\% \\
COURTLISTENER\_CITATION\_LOOKUP & 30.3\% & 25.2\% & 33.3\% & 48.2\% & 34.8\% \\
EDIT\_SCRATCHPAD & 0.2\% & 0.0\% & 0.0\% & 0.3\% & 1.0\% \\
OPEN\_COURTLISTENER\_SEARCH & 7.7\% & 8.1\% & 10.7\% & 10.8\% & 6.7\% \\
OPEN\_WEB\_SEARCH & 3.4\% & 8.3\% & 6.1\% & 4.2\% & 1.7\% \\
READ\_DOCUMENT & 1.4\% & 1.6\% & 2.9\% & 3.7\% & 1.5\% \\
SEARCH\_LOCAL\_OPINION & 31.1\% & 37.3\% & 29.3\% & 14.9\% & 28.6\% \\
THINK & 1.2\% & 0.0\% & 0.1\% & 0.9\% & 0.9\% \\
\hline
\textbf{Average \# of Tool Calls} & 9.7 & 15.3 & 16.8 & 7.5 & 9.3 \\
\hline
\end{tabular}
\caption{\textbf{Action type frequency by model.} GPT-OSS-120B takes the highest average number of steps on each data point. This is likely due to it having the highest percentage of duplicate citation lookups (See Table \ref{tab:redundant_searches}).}
\label{tab:action_type_frequency}
\end{table}

\subsection{Error Analysis}
\paragraph{Handling citations not on CourtListener.} In the test set, there are 132 citations not found on CourtListener. As described in Section \ref{sec:dataset}, these citations were not altered during hallucination injection and are therefore all correct. Any model that flags them as hallucinated produces a false positive.
Table \ref{tab:courtlistener_fp_rate} shows the false positive rate on these citations across models. The variation is substantial: Gemini 2.5 Flash flagged 65.9\% of them as hallucinated, while GPT-5 flagged only 25.0\%. This gap reflects a meaningful difference in verification strategy. Weaker models appear to treat absence from CourtListener as evidence of hallucination, which is a brittle heuristic given that roughly 10\% of real citations in our dataset are Westlaw or Lexis citations that CourtListener cannot resolve.

GPT-5's handling of these cases is more sophisticated. Of the 132 distinct citations missing from CourtListener, it correctly withheld a hallucination judgment on 88. Of these, 40 were ultimately verified through alternative sources: open CourtListener searches occasionally returned case dockets that confirmed a citation's existence even without resolving the Westlaw or Lexis identifier, and in some cases GPT-5 located opinions that cited the same reference. A further 33 citations were left in a pending state, meaning that the agent searched but could not reach a conclusive determination. The remaining cases were never checked, typically due to unusual citation formats or non-standard sources such as FERC citations.

Claude Code agent achieves the lowest false positive rate and it handles these citations differently. Among citations that return a not found status on CourtListener lookup, many of them use Westlaw or Lexis identifiers. CourtListener may in fact have these cases but do not have the Westlaw and Lexis identifiers. Claude Code agent attempts to look up the content of the cited opinion directly to locate the cases instead. However, this approach comes with the caveat that the citation identifiers themselves are not verified, only the content of the case is. 

This pattern illustrates an important trait of effective verification agents: the ability to recognize the limits of a single database, seek corroborating evidence across multiple sources, and calibrate confidence accordingly rather than defaulting to a false positive.

\paragraph{Duplicate actions that do not generate new information.} In many episodes, the agents take near-duplicate actions that do not generate new information. We classify duplicate actions into three categories:

\begin{enumerate}
    \item Duplicate citation lookup: COURTLISTENER\_CITATION\_LOOKUP is called with the same citation string (e.g. "708 F.3d 704") more than once in the same episode.
  \item Duplicate opinion search: SEARCH\_LOCAL\_OPINION is called with the exact same (opinion\_id, query) pair more than once.
  \item Re-search after hit: SEARCH\_LOCAL\_OPINION returns found=True for an opinion, but the model later calls
  SEARCH\_LOCAL\_OPINION on the same opinion X again with a different query that is $>$ 50\% similar to the one that already succeeded. 
\end{enumerate}

Duplicate citation lookup and duplicate opinion search mainly happen when the agent reaches a dead end with its queries. However, instead of recognizing that the citation or the information it is verifying is hallucinated, it attempts to restart to find different information. Re-search after hit type of duplicate happens when the agent tries to confirm a quote or a holding repeatedly. When a follow-up search does not return a result due to changed wording, the agent's belief on the citation becomes unstable even if a previous search returned a match and should have confirmed its correctness. 

\begin{table}[t]
\centering
\begin{tabular}{lcc} 
  \hline
  Model & \# Flagged & False Positive Rate (\%) \\
  \hline
  Gemini 2.5 Flash & 87 & 65.9 \\
  GPT-5 & 33 & \textbf{25.0} \\
  GPT-OSS 120B & 54 & 40.9 \\
  Qwen3-8B & 61 & 46.2 \\
  Qwen3.6-27B & 82 & 62.1 \\
  Claude Code with Opus 4.8 & 14 & \textbf{10.9} \\
  \hline
  \end{tabular}
\caption{\textbf{False positive rate on citations not found on CourtListener.} Calude Code has the lowest false positive rate overall on these citations and GPT-5 has the lowest false positive rate among BOED agents. When a citation is not found on CourtListener, more powerful agents usually attempt to find it from other sources.}
\label{tab:courtlistener_fp_rate}
\end{table}

\begin{table}[t]
\footnotesize
\centering
\begin{tabular}{lcccc}
\hline
Model  & \makecell{Duplicate Citation\\Lookup (\%)} & \makecell{Duplicate Opinion\\Search (\%)} & \makecell{Re-search\\After Hit (\%)} & \makecell{All (\%)} \\
\hline
Gemini 2.5 Flash & \textbf{13.1} & \textbf{2.3} & 0.3 & \textbf{14.1} \\
GPT-5 & 39.7 & 26.4 & 8.2 & 52.3 \\
GPT-OSS 120B & 57.4 & 32.6 & 7.7 & 62.1 \\
Qwen3-8B & 34.4 & 5.4 & \textbf{0.0} & 36.2 \\
Qwen3.6-27B & 20.8 & 10.0 & 1.5 & 23.8 \\
\hline
\end{tabular}
\caption{\textbf{Percentage of episodes containing each type of redundant tool call.} Many models perform duplicate calls that do not generate new information.}
\label{tab:redundant_searches}
\end{table}
\section{Case Studies of Agent Trajectory}
\label{app:trajectory}
In this section, we show a few agent trajectories where the agent is able to leverage information on the internet to verify citations and their contents. 

\paragraph{Verifying citation validity using external sources.} In the first example, the agent looked up a citation, "2006 WL 644442", that is not available on CourtListener. Instead of concluding that the citation is hallucinated after getting no result from CourtListener, the agent searched up the citation on the Internet. It found a federal district-court filing that quotes the citation. The agent then tried to search up the citation on Google Scholar, but it returned no result. In the final belief, the agent correctly recognized that the opinion not being on CourtListener nor Google Scholar does not prove that the citation is hallucinated. 

\begin{tcolorbox}[colback=floralpink!10, colframe=floralpink, title=GPT-5 agent trajectory 1, breakable]
\footnotesize
TRAJECTORY: benjamin-v-trustees-barton-comm-college\_301666810\_part7.pdf \\
Model: GPT-5   \\
Total Steps: 30
\\
Brief excerpt contains: "Walters v. Benedict College, No. 3:04-0952-JFA, 2006 WL 644442 (D.S.C. Mar. 10, 2006)"
\\\\
================================================================================
\\\\
STEP 2 — COURTLISTENER\_CITATION\_LOOKUP
  Cite:   2006 WL 644442\\
  Result: NOT FOUND (404) — "Citation not found: '2006 WL 644442'"\\
\\
TASK BELIEFS (entering step 17): \\
    • Walters v. Benedict College, No. 3:04-0952-JFA, 2006 WL 644442 (D.S.C. Mar. 10, 2006) (unpublished) \u2014 citation existence/consistency. Status: pending (initial web search did not return a direct hit; further verification needed).
\\\\
STEP 3 — OPEN\_WEB\_SEARCH \\
  Query:  Walters v. Benedict College 2006 D.S.C. 3:04-0952-JFA 2006 WL 644442 \\
  Results:
  
    [1] "in the united states district court"\\
        URL:     https://www.govinfo.gov/content/pkg/USCOURTS-paed-2\_14-cv-03191/pdf/USCOURTS-paed-2\_14-cv-03191-0.pdf\\
        Snippet: "Walters v. Benedict College, No. 3:04-0952-JFA, 2006 WL 644442, ov's wrongful termination claim will be dismissed."
\\
TASK BELIEFS (final): \\
    • Walters v. Benedict College, No. 3:04-0952-JFA, 2006 WL 644442 (D.S.C. Mar. 10, 2006) — citation existence/consistency. Status: verified as correct. Note: Previously tied to CourtListener docket ID 4647688; new observation (0 Google Scholar results for the WL cite) does not contradict existence and is consistent with an unpublished, WL-only district court order not indexed on Scholar.
\end{tcolorbox}

\paragraph{Verifying citation content using external sources.}
The second example illustrates the agent leveraging web search to confirm a pincite location when the CourtListener opinion did not contain page number information. After pulling the opinion from CourtListener, the agent was able to verify the content of the quotes from the opinions but pincite correctness remained pending because the opinion did not contain page number. At step 28, the web search returned page-marked secondary sources, so the agent ultimately did not flag the pincite as incorrect. 

\begin{tcolorbox}[colback=floralpink!10, colframe=floralpink, title=GPT-5 agent trajectory 2, breakable]
\footnotesize
TRAJECTORY: kevin-mccarthy-v-patricia-fuller\_238053664\_part12.pdf\\
Model: GPT-5  \\
Total Steps: 30\\ 
Brief excerpt contains: "There is no fear, then, that granting immediate review here will ''swallow the final judgment rule'' or impose significant ''institutional costs.'' Mohawk, 130 S. Ct. at 608-09."
\\\\
================================================================================
 \\\\
  STEP 2 — COURTLISTENER\_CITATION\_LOOKUP
    Cite:   130 S. Ct. at 608-09 \\
    Result: NOT FOUND (404) — "Citation not found: '130 S. Ct. 608'"
  \\\\
    TASK BELIEFS (after step 2): \\
      (no Mohawk belief recorded yet)
  \\\\
  STEP 3 — COURTLISTENER\_CITATION\_LOOKUP
    Cite:   130 S. Ct. 599 \\
    Result: FOUND (200) — 1 cluster: id 1751,
            /opinion/1751/mohawk-industries-inc-v-carpenter/
  \\\\
    TASK BELIEFS (after step 3): \\
      1) Citation: Mohawk, 130 S. Ct. at 608-09 — PENDING. New evidence: CL
         lookup for "130 S. Ct. 608" returned 404 (no match). This raises doubt
         about the S. Ct. pincite as written or reflects a lookup limitation;
         the intended case is likely Mohawk Industries, Inc. v. Carpenter,
         558 U.S. 100, 130 S. Ct. 599 (2009). Need to fetch the Mohawk opinion
         and verify whether pages 608-09 support the cited phrases. \\
      2) Quote: "swallow the final judgment rule" (attributed to Mohawk,
         130 S. Ct. at 608-09) — PENDING. Verbatim appearance, pincite accuracy,
         and context outstanding; uncertainty increased by the 404 on
         130 S. Ct. 608. \\
      3) Quote: "institutional costs" (attributed to Mohawk,
         130 S. Ct. at 608-09) — PENDING. Same verification needs; still
         unverified given the failed S. Ct. pincite lookup.
  \\\\
  STEP 4 — ACCESS\_COURTLISTENER\_OPINION
    Opinion ID: 1751 (Mohawk Industries, Inc. v. Carpenter) \\
    Result: Fetched full opinion. Snippet: "(Slip Opinion) OCTOBER TERM, 2009
            1 Syllabus ..."
  \\\\
    TASK BELIEFS (after step 4): \\
      1) Citation: Mohawk, 130 S. Ct. at 608-09 — PENDING. New evidence verifies
         the case and reporter: CL confirms Mohawk Industries, Inc. v. Carpenter
         at 130 S. Ct. 599 (and 558 U.S. 100). Case existence and name/reporter
         consistency established. Still need to verify pages 608-09 contain the
         cited propositions/phrases. (Associated cluster id: 1751) \\
      2) Quote: "swallow the final judgment rule" — PENDING. Requires verbatim
         match, pincite accuracy to pages 608-09, and contextual accuracy. \\
      3) Quote: "institutional costs" — PENDING. Requires verbatim match,
         correct pincite, and contextual accuracy.
  \\\\
  STEP 28 — OPEN\_WEB\_SEARCH
    Query:  Mohawk Industries v. Carpenter 130 S. Ct. 599 608 609
            "swallow the final judgment rule" "institutional costs" \\
    Results:

      [1] "MOHAWK INDUSTRIES, INC. v. CARPENTER"
          URL:     https://www.law.cornell.edu/supct/html/08-678.ZC.html
          Snippet: "... swallow' " the final judgment rule. Ante, at 5 (quoting
                   Digital ... institutional costs." Ante, at 11.* The Court
                   responds that ..."

      [2] "ADVISORY COMMITTEE ON APPELLATE RULES"
          \\URL:     https://www.uscourts.gov/sites/default/files/fr\_import/AP2010-04.pdf
          Snippet: "swallow' " the final judgment rule. Ante, at 605. (quoting
                   Digital ... the likely institutional costs." Ante, at 608 ...
                   The Court ..."
  \\\\
    TASK BELIEFS (after step 28): \\
      1) Mohawk Industries, Inc. v. Carpenter, 130 S. Ct. 599; 558 U.S. 100
         (associated opinion\_id=1751) — VERIFIED AS CORRECT for case existence
         and name-citation match. \\
      2) Pincite/locator: 130 S. Ct. at 608-09 — PENDING. Update: the stored
         opinion (opinion\_id=1751) has no page markers, so the exact S. Ct.
         pincite cannot be confirmed from this source; do not treat the pincite
         as wrong on this basis. A page-marked source (official U.S., S. Ct., or
         PDF with pagination) is needed for confirmation. \\
      3) Quote: "institutional costs" — VERIFIED AS CORRECT (opinion\_id=1751);
         context aligns with caution against expanding collateral-order appeals. \\
      4) Quote: "swallow the final judgment rule" — VERIFIED AS CORRECT
         (opinion\_id=1751); context aligns with the brief.
  \\\\
\end{tcolorbox}
\section{Model Prompts}
\label{app:model_prompts}
We show the following types of prompts: 1) prompts used to generate legal filings in the motivating experiment, 2) prompts used to extract quotes and holdings from scraped briefs, 3) prompts used to alter quotes and holdings to create hallucinations, and 4) prompts used in BOED agent.

\subsection{Dataset Construction Model Prompts}
\label{app:dataset_prompt}
\begin{tcolorbox}[colback=oceanblue!10, colframe=oceanblue, title=System prompt for extracting case name, breakable]
\footnotesize
    You are a helpful legal assistant. \\
    Please extract the associated case name from the document excerpt based on the reporter citation and CourtListener case name.\\
    The case name associated with the reporter citation may differ from the CourtListener case name, but please ensure it matches exactly how it appears in the document excerpt.\\
    Please follow the input-output format exactly as specified.
\\\\
    TASK:\\
    - Input: \\
        \{\\
            "document\_excerpt": str,   \# An excerpt from a legal brief document \\
            "reporter\_citation": str,   \# A reporter citation found in the excerpt \\
            "case\_name\_courtlistener": str           \# The case name corresponding to the reporter citation from CourtListener \\
        \}\\
    - Output: str,          \# The case name found in the excerpt for this reporter citation, or null if not found
    \\\\
    RULES:
    1. The case name should match exactly how it appears in the document excerpt. It may not match the CourtListener case name.\\
    2. There may be multiple case names in the excerpt; please select the one that best matches the reporter citation.\\
    3. If there is no matching case name in the excerpt, return "None".
\\\\
    
    EXAMPLES:\\
    Example 1:\\
    - Input: \\
    \{\\
        "document\_excerpt": 'in his closing argument constituted a judicial admission of the referral issue that now justifies a new trial. See [D.E. 657-1 at 11-12]. Plaintiffs, however, did not object to the instruction or act in any way on the claimed judicial admission. In any event, the statement in question is not a judicial admission, does not require a new trial, and does no injustice to plaintiffs. The scope of a judicial admission by counsel is restricted to clear and unequivocal statements as to matters of fact that otherwise would require evidentiary proof. See FOP Lodge No. 89 v. Prince George’s County, 608 F.3d 183 , 190 (4th Cir. 2010) (although a lawyer’s statements may constitute a binding admission of a party, any such statement must be “deliberate, clear, and unambiguous” before it will be afforded preclusive effect) (internal citation omitted); Robinson v. McNeil Consumer Healthcare, 671 F. Supp. 2d 975 (N.D. Ill. 2009) (defense counsel’s statement at closing did not qualify as a judicial admission because it was not an unequivocal admission or deliberate waiver, and the jury’s finding of contributory negligence was supported by evidence). Here, the statement was by no means a “deliberate, clear and unambiguous” concession of fact. To begin with,',\\
        "reporter\_citation": '608 F.3d 183',\\
        "case\_name\_courtlistener": 'Fraternal Order of Police Lodge No. 89 v. Prince George’s County'\\
    \}\\
    - Output: 'FOP Lodge No. 89 v. Prince George’s County'
\\\\
    Example 2:\\
    - Input: \\
    \{\\
        "document\_excerpt": 'and class-certification, the District Court concluded that Plaintiff-Appellants’ Motion for Pre-Certification Discovery was moot. Id. This conclusion, based as it was upon an errant ruling as to ascertainability, was equally an error by the District Court. If the Plaintiff-Appellants have not, as yet, established a showing sufficient for class certification, then pre-certification discovery is appropriate. ARGUMENT I. The District Court Erred in Denying the Motion for Class Certification A. Standard of Review The Third Circuit has held that "[i]n reviewing a district court’s judgment on class certification, we apply the abuse of discretion standard." In re Hydrogen Peroxide Antitrust Litig., 552 F.3d 305 , 320 (3d Cir. 2008), as amended (Jan. 16, 2009). A district court will be found to have abused its discretion if its decision "rests upon a clearly erroneous finding of fact, an errant conclusion of law or an improper application of law to fact." *Newton v. Merrill Lynch, Pierce, Fenner \& Smith, Inc.*, 259 F.3d 154, 165 (3d Cir. 2001), as amended (Oct. 16, 2001) (quoting *In re General Motors Corp. Pick Up Truck Fuel Tank Prods. Liab. Litig.*, 55 F.3d 768, 783 (3d Cir. 1995)). **B. Requirements of Class Certification** Federal Rule of Civil Procedure 23(a) allows for'\\
        "reporter\_citation": '552 F.3d 305',\\
        "case\_name\_courtlistener": 'In Re Hydrogen Peroxide Antitrust Litigation'\\
    \}\\\\
    - Output: 'In re Hydrogen Peroxide Antitrust Litig.'
    \\\\
    Example 3:\\
    - Input: \\
    \{
        "document\_excerpt": 'of Pol., 440 F.3d 579 (2d Cir. 2006) ..................................................................4, 6 Grp. Against Smog \& Pollution, Inc. v. Shenago, Inc., 810 F.3d 116 (3d Cir. 2016) ....................................................................17 Ill. Nat’l Ins. Co. v. Wyndham Worldwide Operations, Inc., 653 F.3d 225 (3d Cir. 2011) .....................................................................39 In re Energy Future Holdings Corp., 842 F.3d 247 (3d Cir. 2016) .....................................................................39 In re Makowka, 754 F.3d 143 (3d Cir. 2014) ....................................................................39 In re Plassein Int’l Corp., 366 B.R. 318 (Bankr. D. Del. 2007), aff’d. 388 B.R. 46 (D. Del. 2008), aff’d 590 F.3d 252 (3d Cir. 2009) ......................................40 In re Wickes Trust, No. Civ. A. 2515-VCS, 2008 WL 4698477 (Del. Ch. Oct. 16, 2008) ...........................................................................39, 42 Kirschenbaum v. 650 Fifth Avenue \& Related Props., 830 F.3d 107 (2d Cir. 2016) ....................................................................6 Medtronic, Inc. v. Lohr, 518 U.S. 470 (1996) ..................................................................................20, 21, 26 Menkes v. Prudential Ins. Co. of Am., 762 F.3d 285 (3d Cir. 2014) ....................................................................28 Odyssey Marine Expl., Inc. v. Unidentified Shipwrecked Vessel, 657 F.3d 1159 (11th Cir. 2011) ................................................................23 Oss Nokalva, Inc. v. European Space Agency, 617 F.3d 756 (3d Cir. 2010) ....................................................................22 Peterson v. Islamic Republic of Iran, 627 F.3d 1117 (9th Cir. 2010) .......................................................... 33 Phillips v. Cty. of Allegheny, 515 F.3d 224 (3d Cir.'\\
        "reporter\_citation": '2008 WL 4698477',\\
        "case\_name\_courtlistener": null\\
    \}\\
    - Output: 'In re Wickes Trust'
\end{tcolorbox}

\begin{tcolorbox}[colback=oceanblue!10, colframe=oceanblue, title=System prompt for extracting quotation, breakable]
\footnotesize
    You are a helpful legal assistant. \\
    Please help extract quotations from this document. \\
    Please follow the input-output format exactly as specified. \\
\\\\
    CLASSIFICATION TASK:\\
    - Input: \\
        \{\\
            "document\_excerpt": str,   \# An excerpt from a legal brief document
            "reporter\_citation": str,   \# A reporter citation found in the excerpt
        \}\\
    - Output: str           \# The quotation associated with the citation in the document, or "None" if no quotation is found
    \\\\

    RULES:\\
    1. There may be multiple quotations, but please only return the quotation that corresponds to the given reporter citation.\\
    2. Please only return a quotation if it is a direct quote from the document excerpt in quotation marks.\\
    3. If there is no quotation corresponding to the reporter citation, return "None".
       \\\\

    EXAMPLES:\\
    Example 1:\\
    - Input: \\
    \{\\
        "document\_excerpt": 'in his closing argument constituted a judicial admission of the referral issue that now justifies a new trial. See [D.E. 657-1 at 11-12]. Plaintiffs, however, did not object to the instruction or act in any way on the claimed judicial admission. In any event, the statement in question is not a judicial admission, does not require a new trial, and does no injustice to plaintiffs. The scope of a judicial admission by counsel is restricted to clear and unequivocal statements as to matters of fact that otherwise would require evidentiary proof. See FOP Lodge No. 89 v. Prince George’s County, 608 F.3d 183 , 190 (4th Cir. 2010) (although a lawyer’s statements may constitute a binding admission of a party, any such statement must be “deliberate, clear, and unambiguous” before it will be afforded preclusive effect) (internal citation omitted); Robinson v. McNeil Consumer Healthcare, 671 F. Supp. 2d 975 (N.D. Ill. 2009) (defense counsel’s statement at closing did not qualify as a judicial admission because it was not an unequivocal admission or deliberate waiver, and the jury’s finding of contributory negligence was supported by evidence). Here, the statement was by no means a “deliberate, clear and unambiguous” concession of fact. To begin with,',
        "reporter\_citation": '608 F.3d 183',
    \}\\
    - Output: 'None'
    \\\\
    Example 2:
    - Input: \\
    \{
        "document\_excerpt": 'and class-certification, the District Court concluded that Plaintiff-Appellants’ Motion for Pre-Certification Discovery was moot. Id. This conclusion, based as it was upon an errant ruling as to ascertainability, was equally an error by the District Court. If the Plaintiff-Appellants have not, as yet, established a showing sufficient for class certification, then pre-certification discovery is appropriate. ARGUMENT I. The District Court Erred in Denying the Motion for Class Certification A. Standard of Review The Third Circuit has held that "[i]n reviewing a district court’s judgment on class certification, we apply the abuse of discretion standard." In re Hydrogen Peroxide Antitrust Litig., 552 F.3d 305 , 320 (3d Cir. 2008), as amended (Jan. 16, 2009). A district court will be found to have abused its discretion if its decision "rests upon a clearly erroneous finding of fact, an errant conclusion of law or an improper application of law to fact." *Newton v. Merrill Lynch, Pierce, Fenner \& Smith, Inc.*, 259 F.3d 154, 165 (3d Cir. 2001), as amended (Oct. 16, 2001) (quoting *In re General Motors Corp. Pick Up Truck Fuel Tank Prods. Liab. Litig.*, 55 F.3d 768, 783 (3d Cir. 1995)). **B. Requirements of Class Certification** Federal Rule of Civil Procedure 23(a) allows for'
        "reporter\_citation": '552 F.3d 305',
    \}\\
    - Output: '[i]n reviewing a district court’s judgment on class certification, we apply the abuse of discretion standard.'
\end{tcolorbox}

\begin{tcolorbox}[colback=oceanblue!10, colframe=oceanblue, title=System prompt for extracting holding, breakable]
\footnotesize
You are a legal-text extraction engine. Your job is NOT to interpret law, but to EXTRACT the sentence in the excerpt that states the legal proposition (rule/standard/test) that the given reporter citation is being used to support.
\\
    You must follow the output JSON format exactly.
\\
    - INPUT (JSON):\\
    \{\\
        "document\_excerpt": str, \\
        "reporter\_citation": str \\
    \}
\\
    - OUTPUT (JSON): \\
    \{
    "holding\_sentence": str|null
    \}
\\
\\
    DEFINITIONS:\\
    - "holding sentence" here means: a sentence in the brief excerpt that states a general legal rule/standard/test (e.g., standard of review, burden, legal test).\\
    - It is usually the proposition that appears immediately before or in the same sentence as the citation.
\\\\
    HARD CONSTRAINTS (must obey):
    1) The holding\_sentence MUST be an exact verbatim substring of document\_excerpt (copy-paste exact).\\
    2) Only consider candidate sentences within this window:
    - the sentence that CONTAINS the reporter\_citation, OR
    - the IMMEDIATELY PRECEDING sentence (ONLY if the citation sentence contains little/no proposition text, e.g., just a citation).
    Do NOT choose sentences farther away.\\
    3) If the citation appears only in a string citation / parenthetical outcome list (e.g., "(issuing writ...); Case, cite (issuing...)"), and there is no nearby rule sentence in the allowed window, output has\_holding=false.\\
    4) If no sentence in the allowed window states a general rule/standard/test, output has\_holding=false.\\
    5) When uncertain, prefer NONE (has\_holding=false).
\\ \\
    SCORING HEURISTICS (use to decide):
    - Strong holding indicators: "must", "requires", "is/are", "entitled to", "reviewed", "reversed only if", "standard of review", "burden of proof", "elements", "test".\\
    - Non-holding indicators: pure case-outcome descriptions ("issuing", "affirming", "reversing", "precluding") without a general rule; parenthetical-only summaries; lists of cases.
\\ \\
    PROCEDURE:
    Step 1: Split document\_excerpt into sentences (roughly; punctuation-based is fine).\\
    Step 2: Identify the sentence index j that contains reporter\_citation.\\
    Step 3: Candidate set = \{j\} plus \{j-1 only if needed per rule \#2\}.\\
    Step 4: Pick the best candidate that states a general rule/standard/test.\\
    Step 5: If none qualifies, return has\_holding=false.
\\\\
    Example 1: \\
    - Input: \\
    \{ \\
        "document\_excerpt": 'about the drug-quantity objection, it would impose the same sentence for the same reasons. This Court has repeatedly affirmed sentences under the harmless-error doctrine in similar circumstances. ARGUMENT AND AUTHORITIES The district court did not err, let alone clearly err, in determining the drug quantity attributable to Davalos. Standard of Review Davalos preserved his drug-quantity challenge by objecting to the PSR's drug-quantity determination. (ROA.215-18.) A district court's calculation of the quantity of drugs involved in an offense is a factual finding that is entitled to considerable deference and will be reversed only if clearly erroneous. See United States v. Betancourt, 422 F.3d 240 , 246 (5th Cir. 2005). This Court will deem the district court's factual findings clearly erroneous only if, based on the entirety of the evidence, it is left with the definite and firm conviction that a mistake has been committed. United States v. Akins, 746 F.3d 590, 609 (5th Cir. 2014). 'Under the clearly erroneous standard, if the district court's account of the evidence is plausible in light of the record viewed in its entirety, the court of appeals may not reverse it even though convinced that had it been sitting as the trier of fact, it would have weighed' \\
        "reporter\_citation": '422 F.3d 240', \\
    \} \\
    - Output: "A district court’s calculation of the quantity of drugs involved in an offense is a factual finding that is entitled to considerable deference and will be reversed only if clearly erroneous."\\
\\
\\
    Example 2: \\
    - Input: \\
    \{ \\
        "document\_excerpt": '(issuing writ of mandamus to preclude deposition of the Vice President's chief of staff); *In re United States*, 197 F.3d 310, 314 (8th Cir. 1999) (issuing writ of mandamus to preclude testimony of Attorney General and Deputy Attorney General); *In re FDIC*, 58 F.3d 1055, 1060 (5th Cir. 1995) (issuing writ of mandamus to preclude testimony of three members of the Board of the FDIC); *Bacon v. Department of Housing and Urban Development*, 757 F.2d 265, 269 (Fed. Cir. 1985) (precluding deposition of the Secretary of the Department of Housing and Urban Development); *United States Board of Parole v. Merhige*, 487 F.2d 25 , 29 (4th Cir. 1973) (issuing writ of mandamus to preclude deposition of members of the Board of Parole). regarding the rescission of DACA and the seeking of legal advice regarding that policy decision. Such a document is plainly deliberative and protected by privilege. Document Tab \#74 (RLIT1879) similarly consists of notes written by the Acting Secretary concerning the implementation of a decision to wind down the DACA policy. The district court offered no explanation of how plaintiffs have met their burden of overcoming the privilege. The court likewise plainly erred in declaring that '[d]efendants have waived attorney-client privilege over' \\
        "reporter\_citation": '487 F.2d 25', \\
    \} \\
    - Output: 'None'
    
\end{tcolorbox}

\begin{tcolorbox}[colback=sagegreen!20, colframe=sagegreen, title=System prompt for altering quotation, breakable]
\footnotesize
    You are a helpful linguistic assistant. \\
    Please help mutate some quotes by swapping one or two words with synonyms and keeping the meaning the same.\\
    Please follow the input-output format exactly as specified and ALWAYS return a string for the mutated quotation.\\
\\\\
    TASK:\\
    Input: str   \# A quotation found in the excerpt\\
    Output: str           \# The mutated quotation with one or two words swapped with synonyms
        \\\\
    EXAMPLES:\\
    Example 1:\\
    Input: 'The Act does not even give the Commission power to fire Board members for violations of other laws that do not relate to the Act, the securities laws, or the Board’s authority.'\\
    Output: 'The Act does not even give the Commission power to fire Board members for infringement of other laws that do not relate to the Act, the securities laws, or the Board’s authority.'
    \\\\
    Example 2:\\
    Input: 'Our interpretation of [12 U.S.C.] 1464 is consistent with the broad policy objectives of FIRREA to strengthen the enforcement powers of federal regulators . . . .'\\
    Output: 'Our interpretation of [12 U.S.C.] 1464 is congruous with the broad policy objectives of FIRREA to buttress the enforcement powers of federal regulators . . . .'
\end{tcolorbox}

\begin{tcolorbox}[colback=sagegreen!20, colframe=sagegreen, title=System prompt for altering holding, breakable]
\footnotesize
    You are given a legal holding excerpted from a judicial opinion or legal brief.\\
    Your task is to rewrite the holding to introduce a hallucinated misrepresentation of the law.\\
    You must follow the input-output format exactly.
\\\\

    - INPUT:\\
    \{\\
        "document\_excerpt": str,        \# An excerpt from a legal brief document\\
        "holding": str,        \# An exact sentence extracted from a legal brief\\
    \}\\
    - Output: str       \# A rewritten version of the holding sentence that misrepresents the cited case
    \\\\
    TASK:\\
    Modify the holding sentence to clearly MISREPRESENT the cited case by altering limiting language.
    If the holding sentence contains a limiting statement, REMOVE it to make the holding BROADER.
    If the holding sentence does NOT contain a limiting statement, INSERT a limiting statement to make the holding NARROWER.
\\\\
    HALLUCINATION STRATEGIES (choose ONE):\\
    1) Standard-of-Review Alteration\\
    Change the standard of review (e.g., de novo → abuse of discretion; clear error → plenary review).\\
    2) Burden or Presumption Reversal\\
    Reverse who bears the burden or flip a presumption (e.g., defendant → plaintiff; government-favorable → defendant-favorable inferences).\\
    3) Doctrinal Misclassification\\
    Assign the claim to an incorrect legal framework or constitutional provision (e.g., Fourth Amendment → Fourteenth Amendment due process).\\
    4) Mandatory vs. Discretionary Shift\\
    Convert a mandatory rule into a discretionary one, or vice versa.\\
    5) Threshold or Legal Test Substitution\\
    Replace the governing legal threshold or test with a materially different one that the cited case does not apply.\\
    6) Scope Expansion or Contraction\\
    Expand a narrow holding into a broad rule, or improperly narrow a categorical rule into an exception-based one.\\
    7) Temporal or Procedural Misplacement\\
    Attribute a rule to the wrong procedural stage (e.g., pleading stage → summary judgment; trial → appellate review).
\\\\
    RULES:\\
    1) Output MUST remain plausible in legal writing style.\\
    2) Output MUST fit into the surrounding context of the document excerpt.\\
    3) Output MUST be incorrect if attributed to the cited authority.\\
    4) Do NOT change party names, facts, or the citation.\\
    5) Do NOT mention that the sentence is altered or misleading.\\
\\\\
    WHEN TO FAIL:\\
    - If the holding sentence does not clearly state a legal rule or standard, or is instead about the fact of a case, output "None".\\
    - If the holding cannot be plausibly modified to create misstatement, output "None".
\\\\
    EXAMPLES:\\
    Example 1:\\
    - Input: \\
    \{\\
        "document\_excerpt": 'Bell Atlantic Corp. v. Twombly, 127 S. Ct. 1955, 1974 (2007).  \"A claim has facial plausibility when the plaintiff pleads factual content that allows the court to draw the reasonable inference that the defendant is liable for the misconduct alleged.\"  Ashcroft v. Iqbal, 129 S. Ct. 1937, 1949 (2009).\\\"The plausibility standard is not akin to a 'probability requirement,' but it asks for more than a sheer possibility that a defendant has acted unlawfully.\"  Id.  \"Where a complaint pleads facts that are 'merely consistent with' a defendant's liability, it 'stops short of the line between possibility and plausibility of \"entitlement to relief\".  Id.  B.  Applicable Law\\\"While it is not clear just when minimal police interference becomes a seizure there can be no question that apprehension by the use of deadly force is a seizure subject to the reasonableness requirement of the Fourth Amendment. "  Tennessee v. Garner, 471 U.S. 1, 7, 105 S. Ct. 1694, 85 L. Ed. 2d 1 (1985) (citation omitted).  Thus, a deadly force complaint under 1983 brought by a free citizen must be analyzed according to Fourth Amendment standards.  See Stroik v. Ponsetti, 35 F. 3d 155, 157 (5th Cir. 1994).  Drain v. Galveston County, 999 F. Supp. 929, 933 (S.D. Tex. 1998).  In individual capacity 1983 suits, the Fifth Circuit applies a heightened pleading standard, which requires Plaintiff to plead facts with particularity, focusing specifically on the conduct of the individual who caused the injury.  See Elliot v. Perez, 751 F. 2d 1472, 1479 (5th Cir. 1985).  "The standard for deciding a Rule 12(c) motion is the same as a Rule 12(b) (6) motion to dismiss".  Guidry v. Am Pub. Life Ins. Co. 512 F. 3d 177, 180 (5th Cir. 2007).  The court must accept all well-pleaded facts in the complaint as true.  Kaiser Aluminum \& Chem Sales, Inc. v. Avondale Shipyards, Inc. 677 F. 2d 1045, 1050 (5th Cir. 1982).  A police officer may not seize an unarmed, non-dangerous suspect by shooting him dead.  Colston v. Banhart, 130 F. 3d 96, 103(5th Cir.)  Deadly force cases pose a unique evidentiary problem because the police officer defendants and their colleagues are often the only surviving eyewitnesses, while the person most likely to contradict their story is dead.  Ludwig v. Anderson, 54 F. d 465, 470 (8th Cir.) fn. 3.  C.  Section 1983\\42 U.S.C. 1983 provides a private right of action for the deprivation of rights, privileges, and immunities secured by the Constitution or laws of the United States.  A complaint under 1983 must allege that the acts complained of occurred under color of state law and that the complaining parties were deprived of rights guaranteed by the Constitution or laws of the United States.  Parratt v. Taylor, 451 U.S. 527, 101 S.Ct. 1908, 1913, 68 L. Ed. 2d 420 (1981), overruled on other grounds, Daniels v. Williams, 474 U.S. 327, 106 S.Ct. 662, 88 L.Ed. 2d 662 (1986)  Piotrowski v. City of Houston, 51 F. 3d 512, 515 (5th Cir. 1995).  A complaint under 1983 must also allege that the constitutional or statutory deprivation was intentional or due to deliberate indifference and not the result of mere negligence.  Farmer v. Brennan, 511 U.S. 825, 114 S. Ct. 1970, 128 L.Ed. 2d 811 (1994).  Plaintiffs suing public officials under 1983 must file short and plain complaints that must be factual and not conclusive.  Schultea v. Wood, 47 F. 3d 1427, 1433 (5th Cir. 1995) (en banc).  D.  Qualified Immunity\\In Scheurer v. Rhodes, 416 U. S. 232, 94 S.Ct. 1683, 40 L.Ed. 2d 90 (1974), the Supreme Court held that if the defendant asserts qualified immunity, the complaint should generally not be dismissed for failure to state a claim because the issue of whether immunity applies is a factual question related to the merits.  Id at 250, 94 S.Ct. at 1693.  Generally, government officials performing discretionary functions have qualified immunity, which shields against civil damages liability, so long as the official's actions are reasonable in view of pre-existing law.  Anderson v. Creighton, 482 U.S. 635, 638, 107 S. Ct. 3034, 97 L. Ed. 2d 523 (1987).  To determine whether defendant is entitled to qualified immunity, we ask: (1) whether the facts, taken in the light most favorable to the party asserting the injury, show that the defendants' conduct violated a constitutional right, and (2) whether the right violated was clearly established at the time of the defendants' alleged conduct.  See, Pearson v. Callahan, 555 U.S. 223, 232, 129 S. Ct. 808, 172 L.Ed. 2d 565 (2009) (citing Saucier v. Katz, 533 U.S. 194, 201, 121 S.Ct. 2151, 150 L.Ed. 2d 272 (2001) overruled in part by Pearsons, 555 U.S. at 236, 129 S.Ct. 808).  While it is often appropriate to answer these two questions sequentially, courts are allowed to exercise their "sound discretion in deciding which of the two prongs of the qualified immunity analysis should be addressed first in light of the circumstances in the particular case at hand. "  Id at 236, 129 S.Ct. 808.  "Clearly established' means that the 'contours of the right must be sufficiently clear that a reasonable official would understand that what he is doing violates that right. "  Thompsons v. Upshur County., 245 F. 3d 447, 456 (5th Cir. 2001) (quoting Anderson, 482 U.S. at 640, 107 S.Ct. 3034).  "The defendant's acts are held to be objectively reasonable unless all reasonable officials in the defendant's circumstances would have then known that the defendant's conduct violated the United States Constitution or the federal statute as alleged by the plaintiff. "  Id (citation omitted).  The "defendant's circumstances" includes facts that the defendant knows.  Id.  "However, because qualified immunity turns only upon the objective reasonableness of the defendant's acts, a particular defendant's subjective state of mind has no bearing on whether that defendant is entitled to qualified immunity"  Id. (citation omitted).  An official is eligible for qualified immunity even if the official violates another's constitutional rights".  Id. (citation omitted).  "The Supreme Court has characterized the [doctrine of qualified immjnity] as protecting a 'all but the plainly incompetent or those who knowingly violate the law. "  Cozzo v. Tangipahoa Parish Council-President Gov't, 279 F. 3d 273, 284 (5th Cir. 2002).  (quoting Malley v. Briggs, 47 U.S. 335, 341, 106 S.Ct. 1092, 89 L.Ed. 2d 271 (1986)).  "Whether an official's conduct was objectively reasonable is a question of law for the court, not a matter of fact for the jury"  Brown v. Callahan, 623 F. 3d 249, 253 (5th Cir. 2010).  CONCLUSION',
        "holding": 'Thus, a deadly force complaint under 1983 brought by a free citizen must be analyzed according to Fourth Amendment standards.',
    \}\\
    - Output: "Thus, a deadly force claim alleging the use of deadly force are governed by the Eighth Amendment’s prohibition on cruel and unusual punishment."
    \\\\
    Example 2:\\
    - Input: \\
    \{\\
        "document\_excerpt": '(Trial Tr. 47-48, Apr. 16, 2012).  Mr. Suggs's incontinence, lack of mobility, and inability to use his arms led to the development of numerous pressure sores on various parts of his body.  These are painful areas that are unable to heal without adequate nutrition and hydration and are painful because loss of skin exposes nerve endings.  (Trial Tr. 23, 32, 35, 39-40, Apr. 12, 2012; Trial Tr. 50-51, Apr. 16, 2012).  The ulcer on Mr. Suggs's hip had to be surgically cleaned and repaired, which required cutting off necrotic tissue, cutting healthy tissue, chiseling through to the bone, and then sewing up the surgery site.\\Mr. Suggs was left in pain and cried and moaned for days following the surgery.  (Trial Tr. 78-81, 82-86, Apr. 12, 2012).  Mr. Suggs's cervical compression gradually caused his diaphragm to become paralyzed, which led to his inability to breathe and necessitated five hospitalizations between 1995 and 2000 for respiratory failure, pneumonia, fecal impaction, and dehydration.  Mr. Suggs's bowel moved into his chest and lung area because the diaphragm was not strong enough to hold.  The hospitalizations became more serious until he required a breathing mask, a ventilator, and a tracheostomy (surgical placement of a breathing tube into his windpipe).  On June 30, 2000, Mr. Suggs died from a pneumothorax (collapsed lung) and respiratory cardiac arrest caused by his paralyzed diaphragm and respiratory failure.  (Trial Tr. 120-139, Apr. 11, 2012).\\Standards of Review  The grant of summary judgment is reviewed de novo.  *Ark Initiative v. Tidwell*, 749 F.3d 1071, 1074 (D.C. Cir. 2014).  The denial of a new trial motion is reviewed for abuse of discretion.  *Daskalea v. District of Columbia*, 227 F.3d 433, 443 (D.C. Cir. 2000).  A new trial is unwarranted if the trial error was harmless.  *See Williams v. United States Elevator Corp.*, 920 F.2d 1019, 1022-23 (D.C. Cir. 1990).  Evidentiary rulings are reviewed for abuse of discretion.  *Huthnance v. District of Columbia*, 722 F.3d 371, 377 (D.C. Cir. 2013).  An erroneous evidentiary ruling will be reversed only if the error affected a party's substantial rights.  *Id.*  The lawfulness of jury instructions is reviewed de novo.  *Conseil Alain Aboudaram, S.A. v. de Groote*, 460 F.3d 46, 52 (D.C. Cir. 2006).  Reversal is warranted only if an error affected the outcome of the district court proceedings.  *Czekalski v. LaHood*, 589 F.3d 449, 453-56 (D.C. Cir. 2009).  The determination how to credit a judgment entered upon a jury verdict against a nonsettling defendant with the proceeds a settling defendant paid to the plaintiff is reviewed de novo.\\*Berg v. Footer*, 673 A.2d 1244, 1247 (D.C. 1996).  An award of attorney's fees is reviewed for abuse of discretion.  *West v. Potter*, 717 F.3d 1030, 1033 (D.C. Cir. 2013).  **Summary of Argument**\\The Court should affirm the judgment, except for the fee award, which should be remanded to correct computational errors.',
        "holding": 'The determination how to credit a judgment entered upon a jury verdict against a nonsettling defendant with the proceeds a settling defendant paid to the plaintiff is reviewed de novo.',
    \}\\
    - Output: 'The determination how to credit a judgment entered upon a jury verdict against a nonsettling defendant with the proceeds a settling defendant paid to the plaintiff is reviewed for abuse of discretion, with substantial deference to the trial court’s allocation of settlement proceeds.'
\end{tcolorbox}

\subsection{BOED Agent Prompts}
\label{app:boed_prompt}
\begin{tcolorbox}[title=BOED agent action selection prompt, breakable]
\footnotesize

--- SYSTEM ---
\\\\
You are an LLM agent taking actions within an information environment to solve a task. Each action you take returns an observation that provides information to help you make an accurate prediction.
\\\\
You use Bayesian Optimal Experimental Design (BOED) for action selection, which focuses on reducing uncertainty about task-specific information.
\\\\
\#\# Task
You are verifying citations in a legal brief for hallucinations. Your uncertainty ($\theta$) is: which citations are hallucinated (non-existent, misquoted, or wrong pincite). You reduce that uncertainty by using search and opinion actions to gather evidence; then you submit your final list of hallucinated citations.
\\\\
\#\#\# Domain / task parameters ($\theta$) \\
\#\#\# Task Parameters ($\theta$)
Task instance-specific information needed to identify hallucinated citations in the brief.
\\\\
Information that directly updates your knowledge about which citations/sentences are hallucinated: \\
- For each citation: whether the case exists, whether the case name matches the case name in the citation, whether the quoted language appears in the opinion, whether the pincite is correct \\
- Evidence from CourtListener searches and opinion fetches that confirms or refutes each citation \\
- Any misattributions, fabricated quotes, or non-existent case citations you discover
\\\\
\#\#\#  Domain knowledge: legal case citations \\
- **Citation format**: \\
    Typical format is Case Name, Volume Reporter Page (e.g., 557 F.2d 170). The part after the comma is the reporter citation; "at" introduces a pincite (specific page).\\
- **Quotation format**:\\
    ONLY words, phrases, and sentences that are inside quotation marks are considered direct quotes.\\
    (i.e. in sentence: The PLRA never takes more than 20 percent of a prisoner's assets not only supports the Courts holding of constitutionality, but was a "critical factor" in it, the quoted text is ONLY critical factor, which should appear in the original opinion, without quotes.)
    The presence of ellipses (...) or bracketed ellipses ([...]) only permits omission of intervening text.\\
    (i.e. in sentence: [W]here Congress explicitly enumerates certain exceptions to a general prohibition), the word [W]here would NOT appear verbatim in the opinion and should be OMITTED from string match searches.)\\
    Return the entire sentence as hallucinated if any visible portion of the quotation does not appear verbatim in the opinion. \\
- **Holding format**: The holding the citation is usually IMMEDIATELY PRECEDES the citation.
\\\\
If the case does not exist or the name does not match, the entire citation is hallucinated—no need to separately verify quotes or holdings for that citation.\\
- **Verification hierarchy**:\\
    If a citation is already determined to be hallucinated (e.g., case does not exist or case name does not match), treat all quotes and holdings under that citation as hallucinated.\\
    In this case, ONLY return the citations that are hallucinated, no need to return the quotes and holdings.
\\\\
\#\# Action Selection Objective\\
Choose the action that maximizes Expected Information Gain (EIG) about $\theta$.
Estimate the EIG of an action by considering how much the observation from this action
will reduce your uncertainty and help you learn information about $\theta$.
\\\\
EIG($\theta$ | action) = I($\theta$; observation | action, history)
\\\\
**Action Selection Process:**\\
1. Consider your candidate actions\\
2. For each action, estimate how much its observation would reduce your uncertainty about $\theta$ (task-specific facts)\\
3. Select the ONE action (action type and parameters) with the highest expected information gain\\
\\\\
**Key principles:**\\
- Consider carefully how each action – both its action type and parameters – will inform your task beliefs ($\theta$)\\
- An action that tells you little new provides low information gain\\
- Prefer actions that resolve the most impactful uncertainties for making an accurate prediction\\
\\\\
\#\# Environment\\
$<$environment\_description$>$
\\\\
\#\# Available Actions\\
$<$create\_selection\_actions\_description(action\_space)$>$
\\\\
\#\# Search Capabilities
\\\\
You have access to:
\\
- **COURTLISTENER\_CITATION\_LOOKUP**: Resolve a reporter citation (e.g., `934 F.3d 53`, `143 S. Ct. 1196`) to canonical case info. Use `cite`.\\
- **OPEN\_COURTLISTENER\_SEARCH**: Search CourtListener by case name or citation text. Use `query`.\\
- **ACCESS\_COURTLISTENER\_OPINION**: Fetch an opinion by `opinion\_id`. Full text is stored locally and registered for reading.\\
- **SEARCH\_LOCAL\_OPINION**: Search within a fetched opinion using `opinion\_id` and `search\_string`.\\
- **READ\_DOCUMENT**: Read a fetched opinion in sections. Use `opinion\_id` = `opinion\_$<$id$>$` (or just the numeric `$<$id$>$`), `start\_line` ($\theta$-indexed), and `num\_lines`. Use this to read the full opinion in chunks when SEARCH\_LOCAL\_OPINION is not enough.\\
- **EDIT\_SCRATCHPAD**: Take notes. Use `operation`: `append` (add to end), `insert` (at `position`), `replace` (at `position`), or `clear`. Use `content` for the text and `position` ($\theta$-indexed) for insert/replace.\\
- **OPEN\_WEB\_SEARCH**: Search the open web for legal information. Use `query`.
\\\\
---\\
\#\#\# Case-law search policy (important)\\
- If a **reporter-style citation** is present, **use COURTLISTENER\_CITATION\_LOOKUP first**.\\
- Use OPEN\_COURTLISTENER\_SEARCH only if citation lookup fails or only a case name is available.\\
- Use OPEN\_WEB\_SEARCH: CourtListener is inconclusive or fails to find the case.\\
\\\\
---\\
\#\#\# COURTLISTENER\_CITATION\_LOOKUP – How to use
- Input only the reporter citation AS IS from the brief.\\
- DO NOT omit 'at' and page numbers following it.\\
- If a match is returned, use the associated `opinion\_id`.\\
\\\\
--- \\
\#\#\# OPEN\_COURTLISTENER\_SEARCH – How to search\\
- Never include pincites or page numbers following "at" (they are pinpoint pages, not identifiers).\\
- Reduce citations to volume + reporter + first page only.\\
- Prefer **quoted reporter citations** (e.g., `"557 F.2d 170"`) or **case name queries**, not both.\\
- Do not combine multiple identifiers in one query.\\
- Expect reporter and punctuation variation (e.g., `F.2d` vs `F2d`).\\
- If no high-confidence match is found, escalate to OPEN\_WEB\_SEARCH.
\\\\
---\\
\#\#\# SEARCH\_LOCAL\_OPINION – How to search\\
- Search for SHORT distinctive substrings and use the returned snippet to verify the quoted language.\\
- Look for key noun phrases.\\
- DO NOT include words that are not in quotation marks and words in square brackets.\\
- Use READ\_DOCUMENT to read the full opinion in sections if string search is inconclusive.\\
\\\\
--- \\
\#\#\# Page numbers (star pagination) – How to check pincites
Opinion text returned by SEARCH\_LOCAL\_OPINION and READ\_DOCUMENT contains markers like `[page 846]`.
A marker means **the printed reporter page 846 begins here**; everything after it is on page 846 until
the next marker. SEARCH\_LOCAL\_OPINION also states the page of each match directly in its match header. \\

- A citation's **pincite** is the page after "at" (e.g. in `Smith v. Jones, 934 F.3d 53, 57 (4th Cir. 2019)`
  or `Smith, 934 F.3d at 57`, the pincite is 57). The cited language must actually appear on that page.\\
- **If the quoted language exists in the opinion but on a different page than the pincite, the citation
  IS hallucinated** — report it. A correct quote attached to the wrong page is still a false citation.
  When this happens, **record the segment as hallucinated in your beliefs** (do not file it away as a
  "verified citation with a minor page issue") and **include it in your final answer**. A wrong pincite
  alone is sufficient to make the segment hallucinated, even if the case name and the quoted words are
  otherwise correct.
- If a match spans a page break, the reported page is a range (e.g. `846-847`); a pincite matching
  either end is correct.\\
- Markers are stripped before searching, so **never put `[page N]` into a `search\_string`**, and a quote
  that runs across a page break will still be found.\\
- Not every opinion has page markers. When the observation says page markers are unavailable
  (`page\_markers\_available: false`), the pincite **cannot** be checked — do not report a pincite error
  for that opinion; judge it on the quoted language alone.
\\\\
--- \\
\#\#\# READ\_DOCUMENT – How to read the opinion\\
- Use the opinion\_id to identify which opinion to read.\\
- Iteratively read the opinion in sections to find the quoted language or holdings.
\\\\
---\\
\#\#\# OPEN\_WEB\_SEARCH – How to search\\
- Do not issue strict multi-field queries (case name + docket + WL citation).\\
- Use loose, normalized case names without punctuation or "v." formatting.\\
- Prefer either case name or reporter citation initially.\\
- Avoid quoting long strings unless retrying after a loose search fails.\\
- Use web search for existence checks, context, or recovery when CourtListener fails.
\\\\
---\\
\#\#\# General verification flow\\
1. Resolve the case (citation lookup → CourtListener search → web fallback).\\
2. Fetch the opinion by `opinion\_id`.\\
3. Verify quoted language with SEARCH\_LOCAL\_OPINION.\\
4. Verify holdings with READ\_DOCUMENT, and EDIT\_SCRATCHPAD.
\\\\
\#\# Action Guidelines\\
- **PROVIDE\_FINAL\_RESPONSE**: Provide the final response to the given task example. Use this when you have completed your exploration and want to provide a final response. \\
- **THINK**: Thinking. Use this to reason about your history and current state, plan your next steps, or analyze information before taking action. Use only briefly to synthesize or transition between actions. Do not use repeatedly for planning; proceed to search actions instead. \\
- **OPEN\_WEB\_SEARCH**: Perform Google search on the open internet (web, news, and Google Scholar). Use this to find current information, news, scholarly articles, and web content that may not be in the closed document index. \\
- **OPEN\_COURTLISTENER\_SEARCH**: Search the CourtListener database for legal cases, opinions, and documents. Use this to find relevant legal information and precedents. \\
- **ACCESS\_COURTLISTENER\_OPINION**: Fetch the full opinion from CourtListener by opinion ID. Use after searching to retrieve the complete opinion text for citation verification. The full opinion is stored locally; use SEARCH\_LOCAL\_OPINION to search within it. The response reports whether the opinion carries printed page markers ([page N]), which are needed to verify a citation's pincite.\\
- **COURTLISTENER\_CITATION\_LOOKUP**: Look up a legal citation (e.g. '934 F.3d 53', '143 S. Ct. 1196') on CourtListener. Returns matching opinion(s) if found.\\
- **SEARCH\_LOCAL\_OPINION**: Search for a string within an opinion already fetched with ACCESS\_COURTLISTENER\_OPINION. Returns a snippet around the match, or None if not found. The snippet includes [page N] markers and each match reports the printed page it appears on, so the citation's pincite can be checked. Page markers are stripped before searching, so quotes spanning a page break still match; do not include [page N] in the search string.\\
- **READ\_DOCUMENT**: Read a portion of a document from search results or available documents. Use this to examine specific parts of legal cases, opinions, or other documents.\\
- **EDIT\_SCRATCHPAD**: Edit the agent's scratchpad to take notes, organize thoughts, or maintain working memory. Use this to keep track of important information during reasoning.
\\\\
**About OPEN\_WEB\_SEARCH:**
- OPEN\_WEB\_SEARCH performs Google search on the open internet (web, news, and Google Scholar). Use this to find current information, news, scholarly articles, and web content.
\\\\
**About COURTLISTENER\_CITATION\_LOOKUP:**\\
- For reporter-style citations (e.g. '965 F.2d 962', '143 S. Ct. 1196'), use **COURTLISTENER\_CITATION\_LOOKUP** first with the `cite` parameter. Use OPEN\_COURTLISTENER\_SEARCH only if citation lookup fails or you have only a case name.
\\\\
\#\# Task-Specific Guidance\\
For this task you must **verify citations** by gathering evidence from external sources. Apply the following when choosing actions:
\\
- **THINK has zero information gain**: The observation from THINK only echoes your thought\\
- **To reduce uncertainty about $\theta$, you must use**: OPEN\_COURTLISTENER\_SEARCH (find cases), OPEN\_WEB\_SEARCH (existence checks), ACCESS\_COURTLISTENER\_OPINION (fetch opinion by ID), SEARCH\_LOCAL\_OPINION (search within a fetched opinion for quoted language). These actions return new evidence; THINK does not.\\
- **Reserve PROVIDE\_FINAL\_RESPONSE** until you have used search/opinion actions to verify ALL citations, or you have exhausted steps.\\
\\\\
**Important**:\\
- Select exactly ONE action with all required parameters\\
- Prefer actions for which you expect the observation to reduce the most impactful uncertainties about $\theta$
\\\\
\#\# Response Format\\
Provide your action selection as a JSON object:
\\\\

\{
  ``action'': \{
    ``action\_type'': ``ACTION\_TYPE'',\\
    ``required\_param1'': ``value1'',\\
    ``optional\_param'': ``value''\\
  \},\\
  ``reasoning'': ``$<$brief explanation for why this action optimizes the objective$>$''\\
\}
\\\\

--- USER ---

\#\#\# Current State

\#\#\# Current Beliefs:
$<$current\_beliefs$>$

\#\#\# Current Observation:
$<$observation.result or 'Initial state'$>$

\#\#\# Recent Actions:
$<$format\_action\_history(history, max\_actions=5) or 'No previous actions'$>$

\#\#\# Task Instance
$<$task\_instance\_description or 'N/A'$>$

\#\# Response Requirements (for PROVIDE\_FINAL\_RESPONSE)
When you choose PROVIDE\_FINAL\_RESPONSE, the "response" field must follow this format exactly:
Provide a **list** of hallucinated citations, case names, quotes, and holdings.
\\\\
**Format:** A JSON array of only the hallucinated segments, e.g. `["string1", "string2"]`.
\\\\
**Completeness (required):** Your list must include **every** citation, quote, or holding that you have labeled as hallucinated in your Current Task Beliefs. Do not omit any segment you believe is hallucinated — the response is evaluated against the full set. If you have N items marked hallucinated in your beliefs, your response must contain exactly those N segments (or the citation alone when sub-items are implied).
\\\\
**Wrong pincites count as hallucinated (required):** If your beliefs note that a quote or holding is correct but appears on a **different page than the cited pincite** (e.g. "the pincite is inaccurate", "the correct location is page X", "does not appear on page N"), that segment IS hallucinated and **must** be in your list — even if you separately marked the case name or the quoted words as verified/correct. Return the citation segment as it appears in the brief (including its pincite page). Do not treat a wrong pincite as a minor issue that can be left off the list.
\\\\
**Important**: If a citation itself is hallucinated, it is assumed that all the quotes and holdings within that citation are hallucinated as well so there is no need to return them separately. If no hallucinations are found, return an empty list: `[]`.
\\\\
**No reasoning in the final answer:** The `response` field must contain **only** the list — no reasoning, explanation, or prose. Put all reasoning, analysis, and explanation in the `reasoning` field, not in `response`. Only return the hallucinated segments: If a holding is hallucinated, return the FULL sentence of the holding.
\\\\
\#\# Task\\
Choose the **single next action** that will maximize expected information gain about:
- **Task parameters ($\theta$)**: Instance-specific information for accurate prediction
\\\\
Consider:\\
1. What you already know about $\theta$ (instance-specific evidence and signals)\\
2. What information would most reduce uncertainty about the correct prediction\\
3. Which action is most likely to provide that information
\\\\
You have **$<$steps\_remaining$>$** steps remaining.
[If steps\_remaining $<=$ 1: "If this is your final step, you must use PROVIDE\_FINAL\_RESPONSE."]
\\\\
Provide your response as the required JSON format.
\end{tcolorbox}

\begin{tcolorbox}[title=BOED agent belief update prompt, breakable]
\footnotesize
--- SYSTEM ---
\\\\
You are an LLM agent taking actions within an information environment to solve a task. Each action you take returns an observation that provides information to help you make an accurate prediction.
\\\\
You use Bayesian Optimal Experimental Design (BOED) for action selection, which focuses on reducing uncertainty about task-specific information.
\\\\
You maintain a Bayesian belief p($\theta$), described in natural language, that is updated based on observations from actions taken in the information environment.
Your task is to maintain a list of citations, quotes, and holdings from the brief, described in words.
Keep a numbered or bulleted list. For each item note: (1) the citation, quote, or holding, (2) status: pending, verified as correct, or hallucinated, (3) the associated opinion id if applicable.
\\\\
\#\# Belief Update Process
Your role is to maintain beliefs about the list of citations, quotes, and holdings from the brief. Think of your beliefs as a distribution over possible states of the world, not a single point estimate.
\\\\
When you receive an observation:
- Update your beliefs about the list of citations, quotes, and holdings from the brief based on any new information that directly informs your prediction for this specific task instance.
\\\\
\#\# Environment
<environment\_description>
\\\\
\#\# Principles
- Be **additive and information-dense**: build upon previous knowledge rather than replacing it
- Preserve prior beliefs unless contradicted by new evidence
- Track multiple hypotheses and interpretations, not just a single narrative
- Note the evidence supporting or contradicting different possibilities
- Focus on instance-specific facts, signals, and multiple possible interpretations
- Acknowledge uncertainty and identify what information would be most valuable next
\\\\
--- USER ---
\\\\
\#\# Previous Beliefs
<previous\_beliefs>
\\\\
\#\# New Observation \\
Action type: $<$action\_type$>$ \\
Action parameters: $<$action\_parameters$>$ \\
Observation: $<$observation$>$ \\
\\\\
\#\# Task
Update your beliefs about the list of citations, quotes, and holdings from the brief.
\\\\
Consider:\\
- What new instance-specific information was revealed?\\
- How does it change your understanding of this task instance?\\
- What key uncertainties remain, and what information would be most valuable to resolve them?
\\\\
\#\# Output Format
Provide your updated beliefs in natural language. Your beliefs should be **additive and information-dense** - build upon your previous knowledge rather than replacing it. Show how your understanding has grown and evolved.
\\\\
Structure your response as:
Task Beliefs: [Your list of citations, quotes, and holdings from the brief, described in words. Keep a numbered or bulleted list. For each item note: (1) the citation, quote, or holding, (2) status: pending, verified as correct, or hallucinated.]
\\\\
\#\# Response Format
Provide your updated beliefs as a JSON object:
\\\\
\{\\
    "task\_beliefs": "$<$your list of citations, quotes, and holdings from the brief, described in words. Keep a numbered or bulleted list. For each item note: (1) the citation, quote, or holding, (2) status: pending, verified as correct, or hallucinated.$>$"\\
\}\\
```
\\\\
**Important**: The value must be a text string (natural language), NOT nested JSON.
\end{tcolorbox}

\end{document}